\documentclass{article}

\usepackage{iclr2026_conference}
\usepackage{times}
\usepackage{amsmath,amssymb}

\usepackage{booktabs,makecell,tabularx,array}
\usepackage{adjustbox}
\newcolumntype{Y}{>{\raggedright\arraybackslash}X}

\usepackage{graphicx}
\usepackage{xcolor}
\usepackage{wrapfig}
\usepackage{caption}
\usepackage{tikz}
\usepackage{pgfplots}
\pgfplotsset{compat=1.18}
\usepgfplotslibrary{groupplots}
\usetikzlibrary{positioning,arrows.meta}

\makeatletter
\newcommand{\setcaptype}[1]{\def\@captype{#1}}
\makeatother

\usepackage{microtype}

\usepackage{url}
\usepackage{hyperref}
\hypersetup{hidelinks,pdfauthor={Anonymous authors}}

\newcommand{\zi}[1]{\textcolor{magenta}{}}

\iclrfinalcopy

\title{KiRoSe: Using Kinematic Data for Efficient Segmentation of Robot Demonstrations}
\author{%
\textbf{Shuangyu Xie$^{1}$, Kaiyuan Chen$^{1}$,
Muhammad Zubair Irshad$^{2}$}\\
\textbf{Richard Cheng$^{2}$, Sergey Zakharov$^{2}$,
Ken Goldberg$^{1}$}\\
$^{1}$University of California, Berkeley
\qquad
$^{2}$Toyota Research Institute
}

\begin{document}
\maketitle

\begin{abstract}

Segmenting robot demonstration data into disjoint temporal and semantic subsegments
is valuable for learning, cleaning, retrieval, and analysis.
Segmentation is often performed manually in a tedious and error-prone
process.  Multimodal Large Language Models (MLLMs) have the potential for
automating this process, but directly ingesting raw multi-camera video
streams into MLLMs can be costly and can quickly exhaust context windows.
In this paper, we introduce \textit{KiRoSe} (\underline{Ki}nematics for \underline{Ro}bot \underline{Se}gmentation), a novel framework that focuses on the companion stream of
robot kinematic data.
Inspired by the principle of classical entropy coding, such as Huffman coding, where frequently occurring symbols are represented more compactly, we propose {Sparse Motion Tokens
(SMTs)},  a discrete kinematic vocabulary designed
to compactly represent recurring, temporally localized motion patterns, which we refer to as \emph{kinematic events}.
SMTs are learned by encoding kinematic trajectories with
temporal-segment supervision and quantizing the resulting features. 
We use increasingly available robot demonstration subtask annotations to supervise an open language model to interpret the resulting token sequences.
Evaluation against human-labeled segmentation shows that KiRoSe can substantially outperform baseline segmentation methods in F1 quality and transfers zero-shot to unseen robot embodiments and ego-centric human motion, while preserving the language model's open-vocabulary knowledge. Beyond segmentation, KiRoSe tracks task progress on an unseen robot and detects when an action is being reversed. (website:  \url{https://berkeleyautomation.github.io/kirose/})

\end{abstract}

\begin{figure}[!ht]
    \centering
    \includegraphics[width=\linewidth]{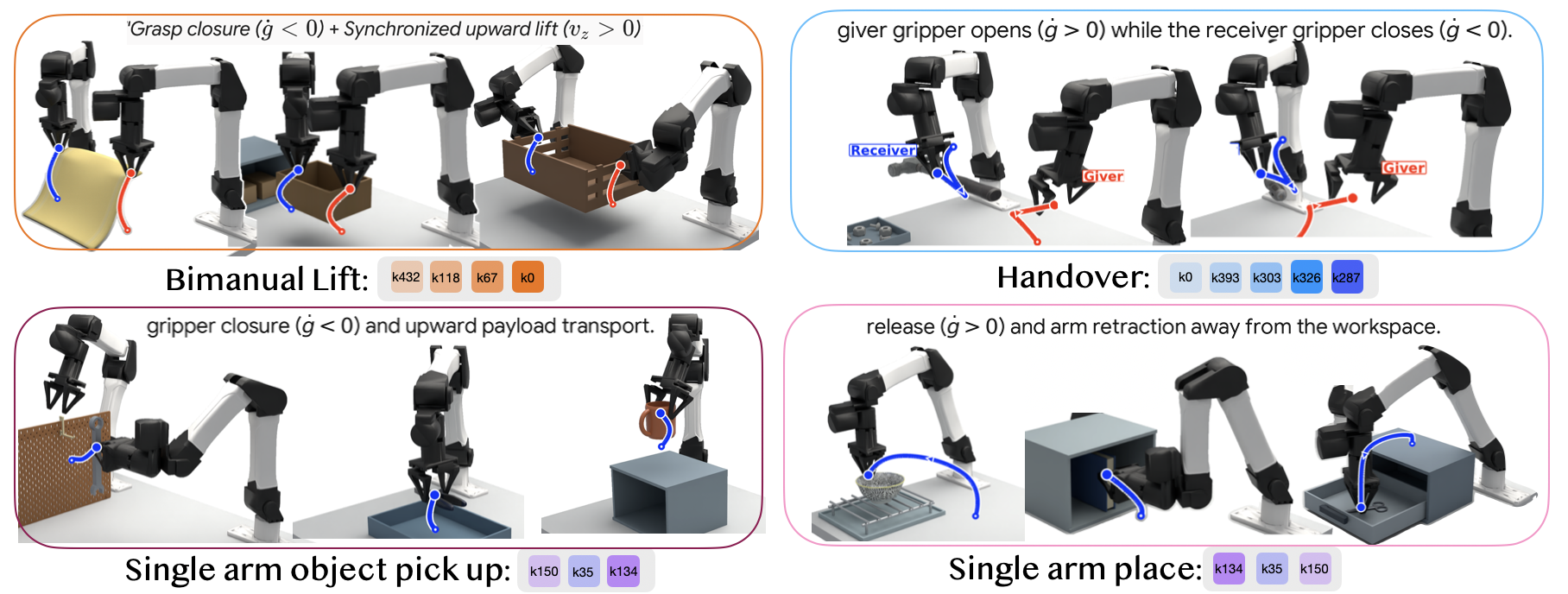}
    \caption{\textbf{Encoding Kinematic Events into Sparse Motion Tokens.} Across diverse manipulation tasks and objects, continuous robot motion decomposes into recurring, temporally localized patterns termed \emph{kinematic events} (e.g., bimanual lift, handover, pick, place). Despite visual and spatial variations, end-effector trajectories within the same motion primitive share structural kinematic signatures. {KiRoSe} compresses kinematic events into a short, ordered sequence of $N$ discrete {Sparse Motion Tokens } (colored blocks; shaded chronologically).}
    \label{fig:teaser}
\end{figure}
\section{Introduction}
\label{sec:introduction}
Large-scale robot demonstration datasets, such as Open X-Embodiment~\citep{oxe2023}, can grow to terabyte or even petabyte scale and span diverse robot embodiments and manipulation tasks. Beyond directly training Vision-Language-Action (VLA) models, understanding the structure within these trajectories is important for effectively using them. Robot demonstrations of “put the cup in the cabinet” and “put the book on the shelf,” for example, share stages such as reaching, grasping, transporting, and placing, potentially mixed with pauses or corrective motions. We refer to recurring, temporally localized motion patterns underlying these stages as \emph{kinematic events} (Figure ~\ref{fig:teaser}). Identifying meaningful kinematic events supports demonstration generation through subtask reuse~\citep{mimicgen2023}, learning and sequencing reusable skills~\citep{skillmimicgen2024}, and retrieving relevant behaviors for policy adaptation~\citep{strap2024}. Such intervals also provide useful units for inspecting demonstration quality, complementing approaches that improve policy learning through data filtering~\citep{demoscore2025} or progress-based weighting~\citep{topreward2026}. 


Manually annotating this structure is tedious and error-prone. Multimodal Large Language Models (MLLMs) offer a promising route to automate subtask annotation, but processing long, synchronized multi-camera videos can incur substantial inference costs and quickly exhaust their context windows. We observe that proprioceptive kinematics provide a compact and information-dense alternative to video: for example, gripper closure may indicate grasping, while a subsequent increase in end-effector velocity may indicate the transition from grasping to object transport (Figure~\ref{fig:teaser}).
However, raw kinematic trajectories are high-frequency and high degree-of-freedom continuous time series without the discrete, semantically structured representation (e.g., directly convert a 1 minute bimanual robot trajectory at 30 Hz consumes approximately 500k text tokens).



In this paper, we introduce \textit{KiRoSe} (\underline{Ki}nematics for \underline{Ro}bot \underline{Se}gmentation), a framework that focuses on using compact kinematics representation to identify action and their boundaries. At its core are Sparse Motion Tokens (SMTs), a low-rate kinematic tokenizer for pretrained VLMs.
Rather than asking a tokenizer to faithfully reconstruct every low-level state, we use the increasingly available subtask annotations in large-scale robot demonstration datasets as supervision for learning a sparse motion vocabulary that preserves information useful for language-model reasoning.
We first encode high-frequency kinematic trajectories into temporally contextual features and quantize them into a compact discrete vocabulary.
To make these tokens interpretable by a language model, we add the SMT vocabulary to a pretrained language model and supervise the reader with trajectory segment annotations containing segment boundaries and natural-language action descriptions.
The tokenizer remains fixed during this stage, so the language model learns to interpret the motion vocabulary rather than reshaping the underlying representation.

We evaluate {KiRoSe} on held-out trajectory segmentation, zero-shot cross-embodiment transfer, semantic description of unseen actions, and task-progress understanding.
Results show that {KiRoSe} achieves a boundary F1 within 1 second of 0.63 vs.\ 0.42 for the strongest alternative kinematic tokenizer on 200 held-out YAM episodes, and, transferred zero-shot to an unseen AgiBot embodiment, a boundary F1 within 0.5 seconds of 0.61 vs.\ 0.30 for GPT-6 Astra.
{KiRoSe} also retains the language model's open-vocabulary knowledge, producing task-appropriate words absent from its training annotations on unseen human tasks; when a subtask is played backward, {KiRoSe} tracks the reversal with a progress correlation of $\rho=-0.68$ on AgiBot, compared with $\rho=+0.88$ for the video-based ReWiND baseline~\citep{rewind}.

To summarize, we make the following contributions: 
(1) \textbf{Sparse Motion Tokens (SMT)}, a boundary-supervised, low-rate kinematic tokenizer that compresses high-frequency robot motion into a compact discrete vocabulary with semantic and temporal motion structure; 
(2) \textbf{KiRoSe}, an end-to-end framework that adapts an open language model to interpret SMT sequences with temporal markers, enabling trajectory segmentation, zero-shot cross-embodiment transfer, open-vocabulary action description, and task-progress understanding;
(3) \textbf{Evaluation Data} suggests substantial performance improvement in segmentation quality over baseline and frontier VLMs.

\section{Related Work}

\paragraph{Trajectory Segmentation for Robot Demonstrations.}
Trajectory segmentation identifies temporal boundaries between motion primitives, with roots in biomechanics and motor control~\citep{sternad1999,viviani1985,morasso1983}. Probabilistic and hierarchical approaches discover these primitives from motion dynamics~\citep{botvinick2009,krishnan2017transition,krishnan2019swirl}, often assuming a smooth, low-dimensional structure~\citep{tanneberg2021skid,iqbal2019fast}. 
Modern robot learning datasets introduce additional challenges through diverse tasks, objects, and embodiments. RoboSegNet~\citep{Chen_2026_CVPR} addresses multi-task trajectory segmentation by fusing visual and kinematic features to predict subtask boundaries. Building on this direction, KiRoSe studies how compact, transition-supervised kinematic tokens can support both temporal boundary prediction and natural-language action descriptions, using the task instruction to contextualize motion.

\paragraph{Kinematics Tokenization.}
Human motion tokenization and its integration with language models have been widely studied. Progress has been made in learning discrete vocabularies through vector quantization and training tokenizers to reconstruct continuous human motion and integrating the resulting tokens with language models for motion generation and understanding~\citep{jiang2023motiongpt,zhou2024avatargpt,wu2025mgmotionllm}. Similar ideas have been explored in robotics, VQ-BeT uses residual vector quantization~\citep{lee2024vqbet}, FAST applies frequency-domain compression~\citep{pertsch2025fast}, and OAT learns ordered tokens supporting prefix reconstruction~\citep{liu2026oat}. These approaches prioritize reconstruction fidelity and efficient generation, whereas KiRoSe uses transition supervision to learn motion tokens that capture the precise kinematic events.

\begin{figure}
    \centering
    \includegraphics[width=\linewidth]{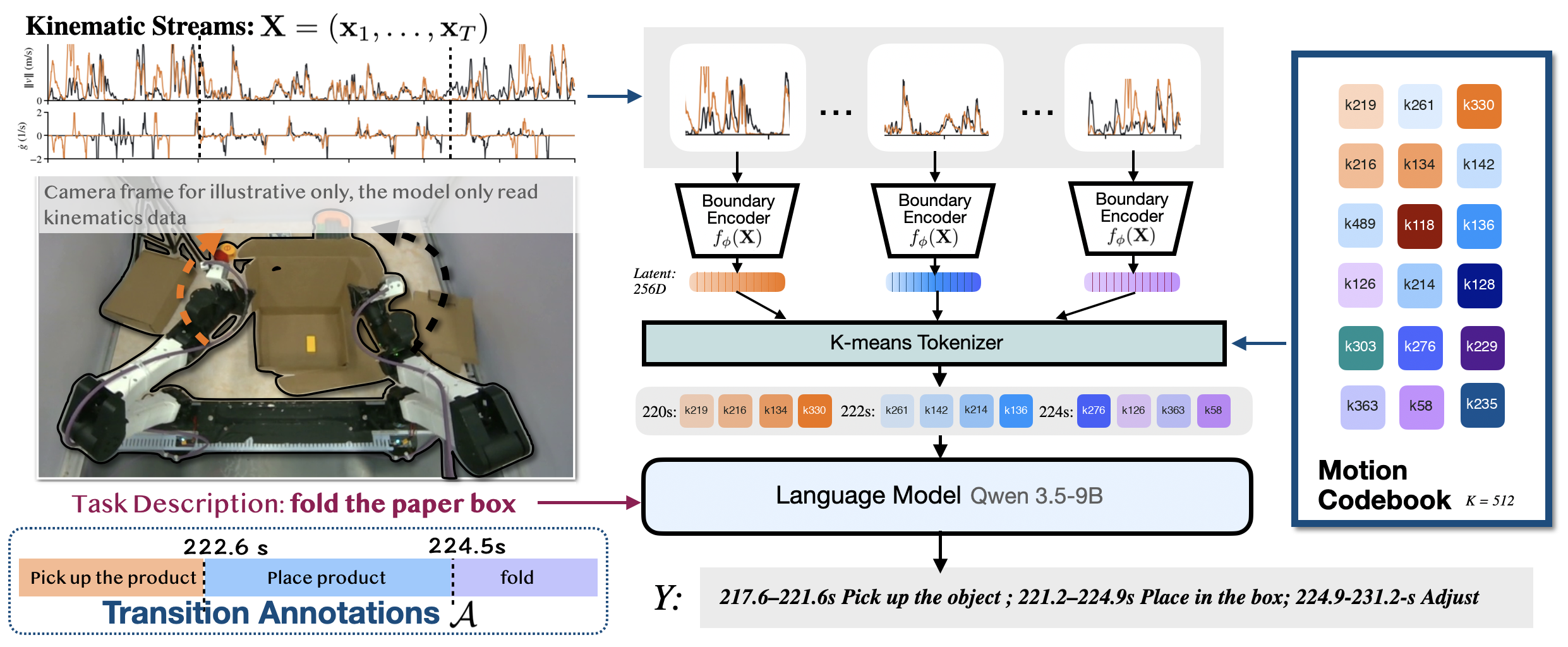}
    \caption{\textbf{KiRoSe Pipeline.}  Given a robot kinematic stream and the task instruction “fold the paper box,” KiRoSe encodes motion using a transition-supervised encoder, then temporally pools and quantizes its features into Sparse Motion Tokens with a learned codebook. The language model reads these tokens, time markers, and the instruction to output action descriptions with start and end times, such as “pick up the object,” “place in the box,” and “adjust.” Transition annotations provide training supervision, while the motion tokenizer remains frozen during language-model adaptation.  }
    \label{fig:overview}
\end{figure}

\section{KiRoSe Pipeline}
\label{sec:method}

KiRoSe comprises two stages (Figure~\ref{fig:overview}). First, we train a kinematic encoder with subtask-transition supervision, then temporally pool and quantize its features into SMTs. Second, we keep the motion tokenizer fixed and adapt a pretrained language model to predict segment boundaries and action descriptions from SMT sequences and the task instruction.

\subsection{Transition Annotation Format}
Let $\mathbf{X} = (\mathbf{x}_1, \ldots, \mathbf{x}_T)$ denote
  a robot kinematic stream, where $\mathbf{x}_t$ is the robot
  state at time step $t$. We annotate action transitions as
  \[
  \mathcal{A} = \{(\tau_i, y_i^{-}, y_i^{+})\}_{i=1}^{N},
  \]
  where $N$ is the number of annotated transitions, $\tau_i$
  is the timestamp of the $i$-th transition in seconds from
  the start of the trajectory, and $y_i^{-}$ and $y_i^{+}$
  are the semantic subtask labels immediately before and
  after the transition, respectively, with
  $y_i^{-} \neq y_i^{+}$. Each annotation therefore marks
  a temporal boundary between two manipulation subtasks (as shown in Figure~\ref{fig:overview}, left corner). Given a kinematic stream $\mathbf{X}$ and its task instruction, trajectory segmentation aims to identify temporal boundaries between subtasks and describe the action performed within each resulting segment.

\subsection{Transition-Supervised Sparse Motion Vocabulary}

\paragraph{Temporal Motion Encoding.}
  \label{sec:temporal_encoding}
Compressing a motion stream into discrete tokens should preserve the temporal structure of manipulation subtasks. We train a temporal encoder to capture motion context relevant to annotated subtask transitions. Denote temporal encoder as $f_\phi$, parameterized by $\phi$, maps $\mathbf{X}$ to a sequence of contextual features:
$
\mathbf{H}=f_\phi(\mathbf{X})
=(\mathbf{h}_1,\ldots,\mathbf{h}_T),
\mathbf{h}_t\in\mathbb{R}^{d},
$
where $d$ is the feature dimension. The encoder uses residual temporal convolutions to aggregate neighboring motion states, allowing each feature $\mathbf{h}_t$ to represent local motion in its temporal context.

During training, we attach an auxiliary boundary-prediction head
$b_\psi:\mathbb{R}^{d}\rightarrow\mathbb{R}$, parameterized by $\psi$, maps each feature to a scalar transition logit. The head predicts a transition probability at each time step:
 $p_t = \sigma\!\left(b_\psi(\mathbf{h}_t)\right),$
where $\sigma$ denotes the sigmoid function. To supervise these predictions, we convert the annotated
transition timestamps $\{\tau_i\}_{i=1}^{N}$ into soft boundary targets $s_t\in[0,1]$, which assign larger values to time steps near annotated transitions. We jointly optimize the encoder and boundary-prediction head using
 $\mathcal{L}_{\mathrm{transition}}(\phi,\psi)
  =
  \frac{1}{T}\sum_{t=1}^{T}
  \ell_{\mathrm{focal}}(p_t,s_t),$
  where $\ell_{\mathrm{focal}}$ is the focal binary
  cross-entropy loss. This objective encourages the motion
  features to retain information useful for locating subtask
  transitions. After training, we discard $b_\psi$ and retain
  $f_\phi$ to produce continuous features for compression into
  discrete motion tokens.

\paragraph{Sparse Motion Tokens.}
  \label{sec:sparse_motion_tokens}
  High-frequency motion features contain substantial temporal
  redundancy. We compress the contextual features
  $\mathbf{H}$ into a low-rate sequence of discrete motion tokens
  through temporal pooling and vector quantization.

  We freeze the temporal encoder $f_\phi$ and apply temporal
  pooling on a uniform grid to obtain a shorter feature sequence
  $\bar{\mathbf{H}}=(\bar{\mathbf{h}}_1,\ldots,\bar{\mathbf{h}}_M)$,
  where $M$ is the number of pooled features and
  $\bar{\mathbf{h}}_j\in\mathbb{R}^{d}$. Each pooled feature is
  then quantized into a discrete motion token.
  We construct a motion vocabulary
  $\mathcal{C}=\{\mathbf{c}_k\}_{k=1}^{K}$, where
  $\mathbf{c}_k\in\mathbb{R}^{d}$ is a codebook centroid and $K$
  is the vocabulary size. Each pooled feature is assigned to its
  nearest centroid, producing a discrete token index
$
  z_j
  =
  \operatorname*{arg\,min}_{k\in\{1,\ldots,K\}}
  \left\|\bar{\mathbf{h}}_j-\mathbf{c}_k\right\|_2^2.
$
  The codebook is fitted using $K$-means on pooled features from
  the training trajectories only, with $K=512$. Specifically,
  letting $\mathcal{P}_{\mathrm{train}}$ denote the collection of
  training features, we minimize the quantization objective
$\mathcal{L}_{\mathrm{quant}}(\mathcal{C})
  =
  \frac{1}{|\mathcal{P}_{\mathrm{train}}|}
  \sum_{\mathbf{v}\in\mathcal{P}_{\mathrm{train}}}
  \min_{k\in\{1,\ldots,K\}}
  \left\|\mathbf{v}-\mathbf{c}_k\right\|_2^2.$
  This objective groups similar contextual motion features into
  a shared discrete vocabulary.

The resulting representation is a discrete motion-token
  sequence $\mathcal{Z}=(z_1,\ldots,z_M)$, with each token
  accompanied by its timestamp on the uniform temporal grid.
  Token positions are independent of annotated or predicted
  boundaries; transition supervision shapes the encoded
  features rather than token locations. Both $f_\phi$ and
  $\mathcal{C}$ remain fixed during downstream language adaptation.

\subsection{Language Model Integration}
  \label{sec:language_reader}

  To recover subtask boundaries and language descriptions from the discrete
  motion representation, we adapt a pretrained language model to
  interpret motion tokens in their temporal and task context. We add $K=512$ motion entries, \texttt{<k0>} through
  \texttt{<k511>}, to the language-model vocabulary, each with a
  trainable embedding. The input combines a task instruction with
  the motion-token sequence $\mathcal{Z}=(z_1,\ldots,z_M)$ defined
  above. We insert textual time markers every second to provide
  a temporal reference for the predicted segment boundaries.

  For supervision, we serialize the reference segmentation as a
  JSON sequence of segments, each containing a start time, an end
  time, and a natural-language description. Segment times are
  expressed in seconds from the start of the trajectory. Let
  $Y=(y_1,\ldots,y_L)$ denote the tokenized JSON response, where
  $L$ is its length and $\ell$ indexes language-model output
  tokens rather than motion time steps. Given the input prompt
  $P$, containing the task instruction, motion tokens, and time
  markers, we optimize
  $
        \mathcal{L}_{\mathrm{read}}
      =
      -\sum_{\ell=1}^{L}
      \log p_\theta\!\left(y_\ell \mid y_{<\ell},P\right),
      \label{eq:reader_loss}
  $
  where $\theta$ denotes the adapted reader parameters and
  $y_{<\ell}$ denotes the preceding response tokens. The loss
  is applied only to response tokens.



\section{Experiments}

We evaluate KiRoSe across trajectory segmentation, cross-embodiment transfer, language generalization, and task-progress understanding. The experiments address four research questions:
(1) Can KiRoSe identify semantic subtask boundaries and descriptions on unseen manipulation trajectories?
(2) Can a motion representation trained on one robot transfer to unseen embodiments without additional training?
(3) Does KiRoSe transfer action semantics to new embodiments, and does adaptation preserve the language model's vocabulary beyond the training annotations?
(4) Does the learned motion representation encode information about task progress, including whether an action is being executed forward or reversed?

\subsection{Implementation and Protocol}

\paragraph{Bimanual-Arm Representation.}
To support transfer across robots and human hands, we use a shared end-effector motion representation:
\begin{equation}
    \mathbf{x}_t=\bigoplus_{h\in{L,R}}
[\mathbf{v}_{h,t}^{\top},\boldsymbol{\omega}_{h,t}^{\top},\dot{g}_{h,t}]^{\top}
\in\mathbb{R}^{14},
\end{equation}
where $\bigoplus$ concatenates left/right linear velocity ($\mathrm{m/s}$), angular velocity ($\mathrm{rad/s}$), and normalized opening rate ($\mathrm{s}^{-1}$). For robots, these signals come from end-effector motion and gripper opening; for humans, from wrist motion, palm orientation, and thumb--index separation.

  




\paragraph{Datasets and Evaluation Splits.}
Table~\ref{tab:xembod} summarizes the training and evaluation splits.
Training uses 5,478 episodes from the YAM bimanual robot in
ABC-130K~\citep{abc130k}, spanning 140 manipulation tasks.
Evaluation is conducted on three splits.
The first contains 200 held-out ABC-130K episodes from the same YAM
embodiment.
Cross-embodiment evaluation uses 2,653 episodes in AgiBot World~\citep{bu2025agibot} and 193 human-hand trajectories in Ego-Exo4D~\citep{grauman2024ego}.
The AgiBot and Ego-Exo4D tasks are not present in the training set.
Some baselines are evaluated on subsets for the cost reason; each table and figure states its episodes.

These splits differ in embodiment, motion characteristics, and temporal structure (Figure~\ref{fig:embodiments}). AgiBot and Ego-Exo4D contain fewer segments and annotated boundaries per minute than ABC-130K. Semantic overlap also decreases: training verbs cover 99\% of held-out ABC-130K segments, compared with 61\% for AgiBot and 35\% for Ego-Exo4D (Table~\ref{tab:xembod}). Evaluation thus tests generalization across episodes, embodiments, task semantics, and temporal structure.


\begin{table}[t]
\centering
\small
\caption{
\textbf{Training and evaluation splits.}
Duration and segments per episode are reported as mean $\pm$ SD.
The final column reports the number of distinct action verbs, with the
percentage of segments whose verb occurs in the ABC-130K training
annotations in parentheses.
}
\label{tab:xembod}

\setlength{\tabcolsep}{2.5pt}
\renewcommand{\arraystretch}{1.1}

\begin{adjustbox}{max width=\linewidth}
\begin{tabular}{@{}lllrrccrrr@{}}
\toprule
Dataset &
Embodiment &
Purpose &
Episodes &
Tasks &
\shortstack{Duration\\(s)} &
\shortstack{Segments\\per ep.} &
\shortstack{Seg. length\\(s)} &
\shortstack{Bnd.\\per min} &
\shortstack{Shared Verbs\\(in train)} \\
\midrule

\multicolumn{10}{@{}l}{\textit{Training}} \\[1pt]

ABC-130K
& YAM
& Training
& 5,478
& 140
& $173.3 \pm 66.5$
& $26.6 \pm 15.0$
& 4.1
& 9.2
& 83 (100\%) \\

\midrule

\multicolumn{10}{@{}l}{\textit{Evaluation}} \\[1pt]

ABC-130K
& YAM 
& Held-out Test
& 200 
& 10
& $154.6 \pm 57.6$ 
& $24.3 \pm 13.4$
& 3.9 
& 9.4 
& 24 (99\%) \\


AgiBot World
& AgiBot G1
& Cross-Robot-Emb.\
& 2,653
& 3
& $94.4 \pm 31.1$
& $10.8 \pm 3.1$
& 6.8
& 6.9
& 12 (61\%) \\

Ego-Exo4D
& Human hands
& \shortstack[l]{Transfer to Human}
& 193
& 12
& $81.4 \pm 53.5$
& $10.1 \pm 5.9$
& 4.8
& 7.4
& 69 (35\%) \\

\bottomrule
\end{tabular}
\end{adjustbox}
\end{table}


\paragraph{Metrics.}
We use complementary metrics to evaluate the temporal and semantic information of robot demonstration semantic segmentation, corresponding to two questions: \emph{when} manipulation actions change and \emph{what} action occurs within each segment. \textit{Boundary F1}, a metric introduced by ~\citep{Chen_2026_CVPR} to measure temporal segmentation accuracy independently of the predicted action description by matching predicted and reference segment boundaries one-to-one within a temporal tolerance and macro-averaging F1 across episodes; multiple tolerances are reported to distinguish precise boundary localization from approximate transition detection. \textit{Grounded F1} evaluates temporal and semantic correctness jointly: predicted and reference segments are first matched one-to-one when their interval IoU is at least $0.5$, and a match is counted as correct only when the corresponding action descriptions also agree. We report two variants: Lexical Grounded-F1 requires normalized description-token F1 $\geq 0.5$, while Embedding Grounded-F1 requires MiniLM cosine similarity $\geq 0.6$. The lexical metric emphasizes agreement in action vocabulary, whereas the embedding metric allows semantically equivalent descriptions with different wording. Control annotations such as \texttt{mistake}, \texttt{pause}, \texttt{wait}, \texttt{idle}, and \texttt{reset} are excluded from action-description scoring.

\subsection{Trajectory Segmentation}

We compare KiRoSe with FAST~\citep{pertsch2025fast}, OAT~\citep{liu2026oat}, and MotionGPT~\citep{jiang2023motiongpt}, pairing each tokenizer with a Qwen3.5-9B~\citep{qwen35} reader adapted using the same LoRA training procedure. To examine the contributions of the learned features and language-model reader, we also evaluate two variants of our method: \emph{Latent + MLP}, which predicts boundaries directly from continuous encoder features, and \emph{SMT codes + MLP}, which uses the quantized representation but replacing language model with a multilayer perception head. All methods are evaluated on 200 unseen YAM episodes from 10 held-out tasks.

We first evaluate whether the motion representation preserves signals that
are informative for semantic action transitions. Even reconstruction-based
kinematic tokenizers such as FAST and OAT improve Boundary F1
over the zero-shot Qwen3.5-9B video baseline, suggesting that kinematics
provide a more direct signal for localizing manipulation transitions than
visual observations alone in this setting. SMT further improves over FAST,
OAT, and MotionGPT across all temporal tolerances
(Table~\ref{tab:main}). Because these tokenizers are paired with the same
Qwen3.5-9B reader and adaptation procedure, the remaining difference is
primarily associated with how the motion signal is represented.
Reconstruction-oriented tokenizers are designed to preserve low-level
trajectory detail, whereas KiRoSe is trained to emphasize changes in motion
that coincide with subtask transitions. Signals such as changes in
end-effector velocity, rotation, and gripper state are therefore represented
in a form more directly aligned with manipulation boundaries, yielding
0.63 Boundary F1 at $\pm$1\,s compared with approximately 0.41--0.42 for
the alternative kinematic tokenizers.

The evaluation data also suggest that the motion representation retains
semantic information beyond transition timing. KiRoSe obtains higher lexical
and embedding Grounded F1 than the alternative kinematic tokenizers,
indicating that predicted segments are more often paired with descriptions
consistent with the corresponding reference actions. Therefore, 
SMT-based transition-supervised motion tokens preserve information about both
{when} manipulation actions change and the motion context needed to
interpret {what} action is being performed.

\begin{table}[t]
\centering
\caption{
\textbf{Demonstration semantic segmentation on 200 YAM episodes from 10 held-out tasks.}
\textbf{Left:} Boundary F1@$\pm$0.5/1/2\,s and Grounded F1 using lexical or embedding-based
matching. FAST, OAT, MotionGPT, and KiRoSe use the same LoRA-adapted
Qwen3.5-9B reader; baseline input and training-set differences are listed in Appendix~\ref{app:tokenizers}.
$^\dagger$Video and task instruction.
\textbf{Right:} Boundary F1@$\pm$1\,s as function of  motion-token rate.
}
\label{tab:main}

\begin{minipage}[c]{0.58\textwidth}
\centering
\footnotesize
\setlength{\tabcolsep}{3pt}
\begin{adjustbox}{max width=\linewidth}
\begin{tabular}{lccccc}
\toprule
 & \multicolumn{3}{c}{Boundary F1} &
   \multicolumn{2}{c}{Grounded F1} \\
\cmidrule(lr){2-4}\cmidrule(lr){5-6}
Method &
$\pm$0.5\,s &
$\pm$1\,s &
$\pm$2\,s &
Lex. &
Emb. \\
\midrule
Qwen3.5-9B zero-shot$^\dagger$
& 0.09 & 0.16 & 0.22 & 0.05 & 0.07 \\
\midrule
FAST
& 0.22 & 0.42 & 0.50 & 0.12 & 0.17 \\
OAT
& 0.22 & 0.41 & 0.51 & 0.14 & 0.20 \\
MotionGPT
& 0.23 & 0.41 & 0.50 & 0.13 & 0.16 \\
\midrule
KiRoSe: Latent + MLP
& 0.44 & 0.52 & 0.56 & -- & -- \\
KiRoSe: SMT codes + MLP
& 0.38 & 0.45 & 0.49 & -- & -- \\
\textbf{KiRoSe}
& \textbf{0.48} & \textbf{0.63} & \textbf{0.71}
& \textbf{0.43} & \textbf{0.49} \\
\bottomrule
\end{tabular}
\end{adjustbox}
\end{minipage}
\hfill
\begin{minipage}[c]{0.40\textwidth}
\centering
\includegraphics[width=\linewidth]{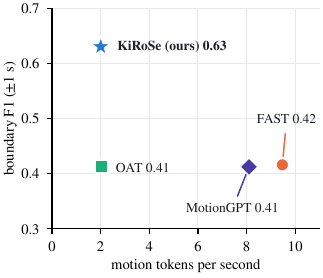}
\end{minipage}
\par\vspace{-0.2in}
\end{table}


\subsection{Cross Embodiment Generalization}

\begin{figure}[t]
\centering
\includegraphics[width=\textwidth]{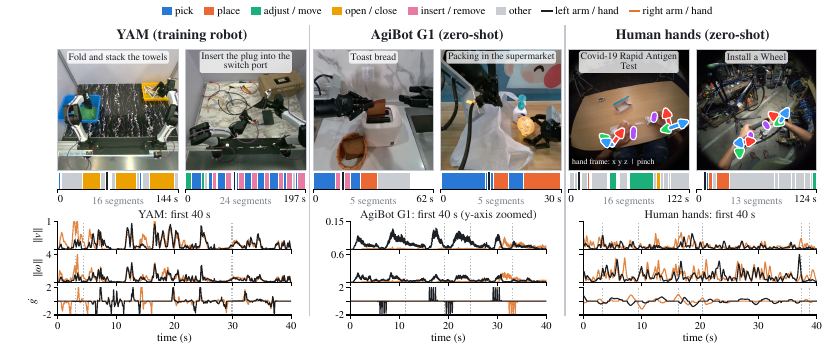}
\caption{ \textbf{Evaluation Dataset.} Top: example tasks and annotated action timelines, with colors indicating action categories. Bottom: representative 40-second traces of linear speed, angular speed, and opening rate for the left (black) and right (orange) arms or hands. The datasets differ in task semantics, temporal structure, and motion characteristics.}
\label{fig:embodiments}
\end{figure}

We evaluate KiRoSe in its generalization capability for demonstrating semantic segmentation  along two dimensions: transfer of transition timing and transfer of action semantics. 
As summarized in
Table~\ref{tab:xembod}, all trainable components are learned exclusively
from ABC-130K trajectories collected with YAM, while AgiBot G1 and
human-hand trajectories are evaluated without target-embodiment training.
Figure~\ref{fig:embodiments} illustrates the resulting domain shift:
the three embodiments share the same end-effector representation but
differ in motion statistics, temporal structure, visual appearance, and
annotation semantics.

\begin{wrapfigure}{r}{0.5\textwidth}
\vspace{-8pt}
\centering
\includegraphics[width=0.49\textwidth]{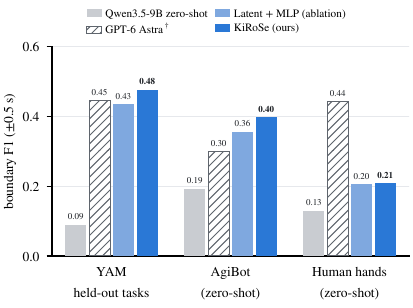}
\caption{Boundary F1 ($\pm$0.5\,s) per embodiment. KiRoSe is trained on YAM only; AgiBot and human hands are
zero-shot. }
\label{fig:xembod}
\vspace{-12pt}
\end{wrapfigure}
\paragraph{Transfer of Transition Timing.}
We first examine whether the temporal structure learned from YAM transfers
to unseen embodiments. Figure~\ref{fig:xembod} reports Boundary F1 to measure \emph{when} the kinematic event happens.

The transfer from YAM to AgiBot involves substantial domain shifts:
the two datasets use different robot embodiments and teleoperation interfaces, exhibit different motion statistics, and differ in both segment duration and annotation granularity (Figure~\ref{fig:embodiments}
and Table~\ref{tab:xembod}). Despite these differences, KiRoSe transfers without any AgiBot training data, obtaining 0.40 Boundary F1 at $\pm0.5$\,s AgiBot testset, compared with 0.48 on held-out YAM trajectories. We ablate with latent + MLP variant, which predicts boundaries directly from the learned motion features, also transfers effectively, achieving 0.36 on AgiBot. The video-based Qwen3.5-9B baseline obtains 0.19 on AgiBot episodes with video. 
We also measure the performance of state-of-the-art vision-language model GPT-6 Astra and the results suggest that KiRoSe outperforms Astra for hold-out set performance on YAM and AgiBot zero-shot transfer. 
The retained boundary localization across these domain
shifts suggests that transition-supervised motion features capture changes in end-effector motion that are shared across robot embodiments, rather than transition patterns specific to YAM. Removing the motion
tokens or replacing them with tokens from another trajectory reduces Boundary F1 on the two robot embodiments, further indicating that the transferred predictions depend on the trajectory-specific motion signal.

\paragraph{Action Semantic Generalization.}
\label{sec:language-generalization}

We next evaluate action semantics on the AgiBot episodes by comparing with GPT-6 Astra, using two Grounded F1 variants. \emph{Action-level} Grounded F1 requires only the verb class to match; \emph{full-description} Grounded F1 also requires the wording, including object names, to match. KiRoSe reaches 0.59 action-level Grounded F1 versus 0.34 for Astra.

\begin{table}
\centering
\caption{\textbf{Unseen-task vocabulary on Ego-Exo4D.}
Supplied words appear in the task's annotations but not in YAM training
or the prompt. Hits: trajectories with a supplied word.}
\vspace{-10pt}
\label{tab:new-word-transfer}
\small
\begin{tabular}{@{}lrrl@{}}
\toprule
Task & Traj. & Hits & Supplied words \\
\midrule
Omelet         & 36 & 17 & pan, egg, mixture, whisk \\
COVID-19 test  & 65 & 14 & swab \\
Tomato \& eggs & 12 & 11 & pan \\
Flat tire      & 13 & 8  & inner (tube), lever \\
Noodles        & 9  & 3  & pot, stir \\
Salad          & 26 & 2  & toss, dressing \\
\bottomrule
\end{tabular}
\end{table}


Human activities in Ego-Exo4D introduce a larger semantic shift: only 35\% of annotated segments use verbs observed during training, compared with 61\% for AgiBot (Table~\ref{tab:xembod}). We therefore complement the robot evaluation with a qualitative analysis on Ego-Exo4D human-hand trajectories from 12 tasks that are absent from YAM training.  YAM-trained KeRoSe receives only the
task name and motion-token sequence.
As reported in Table~\ref{tab:new-word-transfer}, supplied
words occur in 6 of the 12 tasks, including ``tube'' and ``lever'' for Flat tire
installation, ``swab'' for a COVID-19 test, and ``whisk'' and ``egg'' for
preparing an omelet. These words are consistent with what a language model can
infer from the task name, so this result suggests that as  LoRA adaptation with KeRoSe preserves the language model's open-vocabulary knowledge, rather than as semantics recovered from the training dataset . 

\subsection{Cross-Embodiment Progress Estimation}

We further hypothesize that transition-supervised motion representations also retain ordered motion patterns \emph{within} a segment. For example, this may indicate whether an action is progressing toward completion or being undone. A useful dense progress signal should increase during successful execution and decrease when an action regresses: ReWiND~\citep{rewind}, for example, uses reversed trajectories to represent decreasing progress. We therefore use action reversal as a controlled test: if a model represents progress rather than elapsed time, reversing an action should reverse its predicted progress.

\begin{figure}[t]
  \centering
  \includegraphics[width=\linewidth]{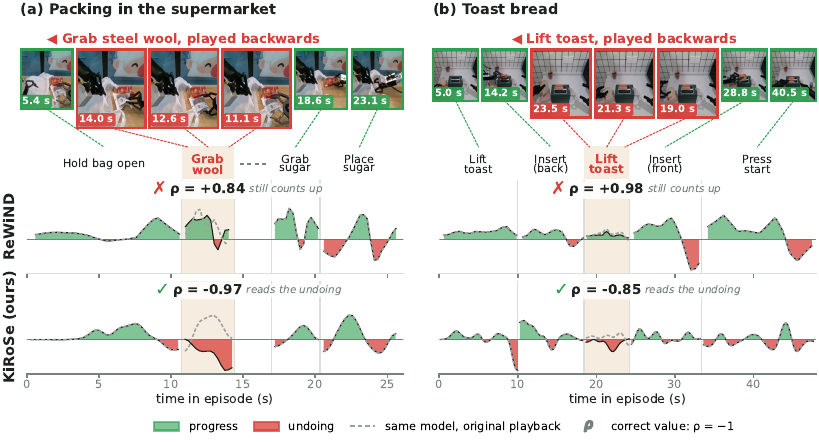}
  \vspace{-0.3em}
  \caption{\textbf{Qualitative reversal probe on AgiBot (zero-shot).} One subtask per episode is
  played backwards in place (red frames; stamps give original recording time). Curves show the
  signed progress rate (green: rising, red: falling); dashed: same model on the original video.
  ReWiND keeps counting up while the task is undone; KiRoSe's progress falls; Table~\ref{tab:reversal}
  reports statistic result.}
  \label{fig:reversal}
\end{figure}

\paragraph{Setup.}  We use 2{,}586 annotated
subtasks of at least 3.2\,s from 357 AgiBot episodes across three tasks. Each subtask is evaluated
as recorded and reversed in place, with the rest of the episode unchanged: video frames are played
backward and kinematic velocities are negated, which corresponds to motion that undoes the action.
We compare with ReWiND~\citep{rewind}, retrained with the authors' public code on MetaWorld (MW)
and on Open-X (OXE), and with the released Robometer-4B~\citep{robometer}; both are trained with
supervision that includes reversed or rewound examples. For KiRoSe, the 2.76M-parameter SMT encoder,
trained only for transition detection on YAM, stays frozen. A 68k-parameter probe reads its features
and the displacement accumulated since the subtask start, and is trained only on YAM subtasks (Appendix~\ref{app:progress}).  We report the Spearman correlation $\rho$
between predicted progress and playback time ($\rho>0$ forward, $\rho<0$ when undone), and the
fraction of reversed subtasks with $\rho<0$. A model that tracks elapsed time stays positive in both
directions.

\paragraph{Results and Analysis.}
\begin{table}[t]
\centering
\caption{\textbf{Zero-shot reversal progress evaluation on AgiBot.}
The same 2,586 subtasks (357 episodes) are evaluated forward and reversed.
``Rev.\ $\rho<0$'' is the percentage of reversed subtasks
with negative progress correlation.}
\label{tab:reversal}
\small
\setlength{\tabcolsep}{4pt}
\begin{tabular}{@{}lccc@{}}
\toprule
Model &
Fwd.\ $\rho\,\uparrow$ &
Rev.\ $\rho\,\downarrow$ &
Rev.\ $\rho<0$ (\%) $\uparrow$ \\
\midrule
ReWiND (MW)  & $\mathbf{+0.88}$ & $+0.88$ & 1\% \\
ReWiND (OXE) & $+0.65$ & $+0.63$ & 4\% \\
Robometer-4B & $+0.71$ & $+0.06$ & 46\% \\
\textbf{KiRoSe} & $+0.51$ & $\mathbf{-0.68}$ & \textbf{86\%} \\
\bottomrule
\end{tabular}
\end{table}
Table~\ref{tab:reversal} compares whether predicted progress follows the
direction of action execution under reversal. Overall, SMT maintains a
positive progress correlation during forward execution and switches to a
negative correlation when the same actions are reversed. In contrast, the
video-based reward models remain positively correlated or approach zero
under reversal, despite being trained with progress-related supervision
that includes reversed or rewound trajectories.
On AgiBot, SMT obtains $\rho=+0.51$ on forward trajectories and $\rho=-0.68$ on
reversed trajectories, with predicted progress falling on 86\% of reversed subtasks. ReWiND remains positively correlated after reversal,
with $\rho=+0.88$ and $+0.63$ for its MetaWorld and Open-X checkpoints,
respectively, while Robometer-4B approaches zero at $\rho=+0.06$.
Because neither the SMT-encoder nor the lightweight head sees AgiBot, this
suggests that transition-supervised kinematic features retain the direction
of within-segment motion in a form that transfers across embodiments.
Reversing an action changes the direction of end-effector velocity and
reverses motion events such as gripper closing and opening, which are also
informative for identifying semantic transitions.

\section{Conclusion}
We presented KiRoSe, a framework for identifying \emph{when} manipulation actions change and \emph{what} action occurs within each segment. KiRoSe uses SMTs to compress high-frequency kinematics into a transition-supervised representation for language-model interpretation.
Evaluation on YAM, AgiBot G1, and Ego-Exo4D indicates that SMT preserves
both temporal and semantic information, transfers transition structure
zero-shot across robot embodiments together with action identity, and
preserves the language model's open-vocabulary knowledge. Beyond segmentation,
the same frozen representation supports within-segment progress estimation
on an unseen robot, including detecting when an action is being undone,
even though the encoder is never trained for progress or reversal. Future work can extend
this representation toward richer contact and object-centric signals, broader
robot embodiments, and downstream demonstration retrieval, data
curation, and reward modeling.

\subsubsection*{AI Use Statement}
We used generative AI tools to assist in revising the kinematic encoder design, including discussing architectural choices and elaborating implementation details. We also used these tools to refine manuscript wording, organization, and figure captions, and to assist with literature searches and LaTeX formatting. All AI-assisted suggestions and content were reviewed by the authors, who take full responsibility for the methodology, experiments, results, and final manuscript.

\bibliography{references}
\bibliographystyle{iclr2027_conference}
\clearpage
\appendix
\newcommand{\appfig}[2]{%
  \IfFileExists{#1}{\includegraphics[width=#2]{#1}}{%
    \fbox{\parbox[c][1.2in][c]{\dimexpr#2-2\fboxsep-2\fboxrule\relax}{%
      \centering\small Missing figure: \texttt{\detokenize{#1}}}}}}

\section{Qualitative Segmentation Examples}
\label{app:examples}

\begin{figure}[p]
  \centering
  \appfig{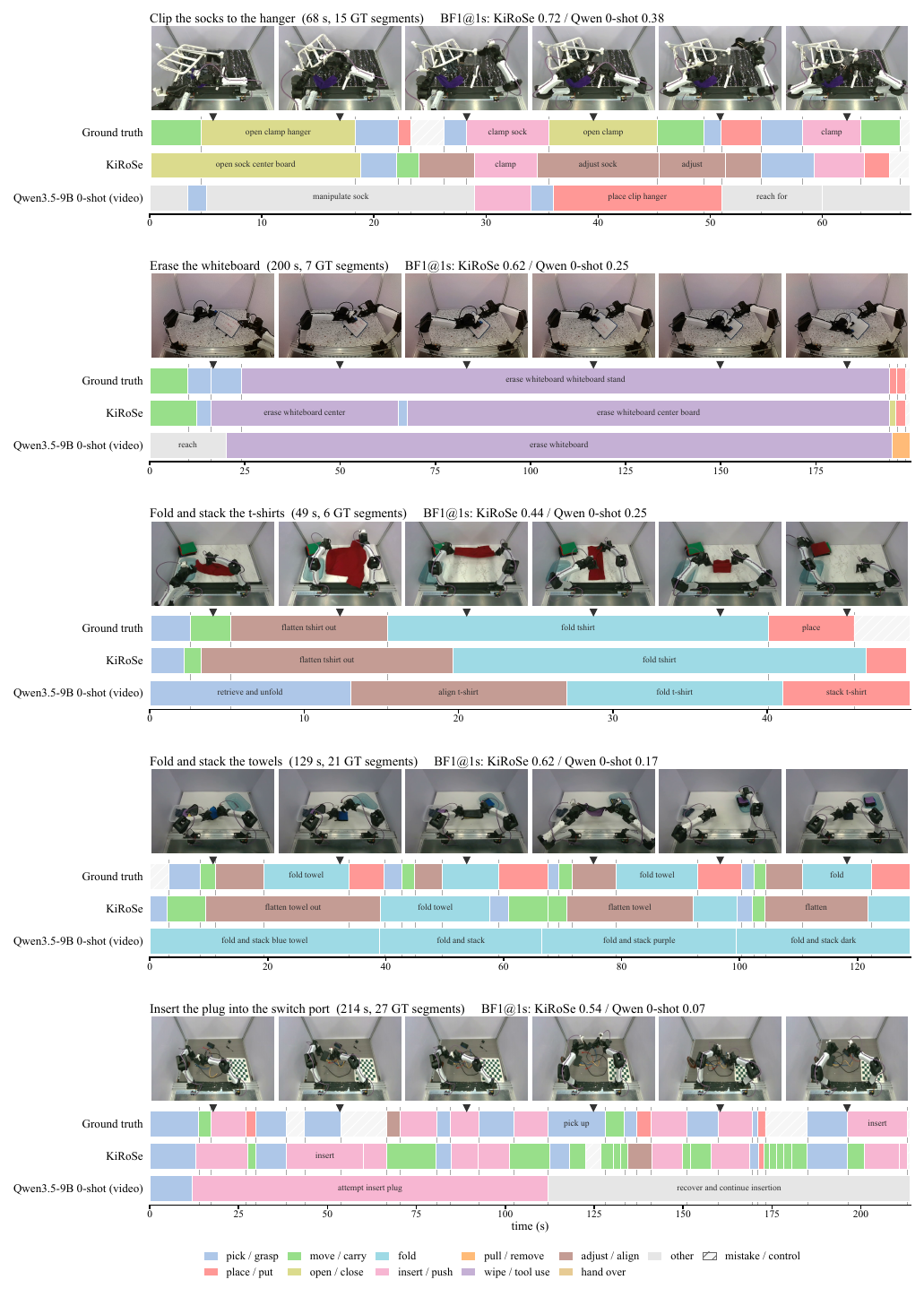}{\linewidth}
  \caption{\textbf{Segmentation examples on held-out YAM tasks (ABC-130K).} The zero-shot video
  model tends to emit a few coarse segments that restate the instruction (``fold and stack blue
  towel'', ``attempt insert plug''), whereas KiRoSe recovers the repeated grasp--place,
  flatten--fold, and insert cycles of the reference from kinematics alone. KiRoSe's Boundary F1 at
  $\pm1$\,s exceeds the video baseline on all five episodes (0.44--0.72 versus 0.07--0.38).}
  \label{fig:app_seg_yam}
\end{figure}

\begin{figure}[p]
  \centering
  \appfig{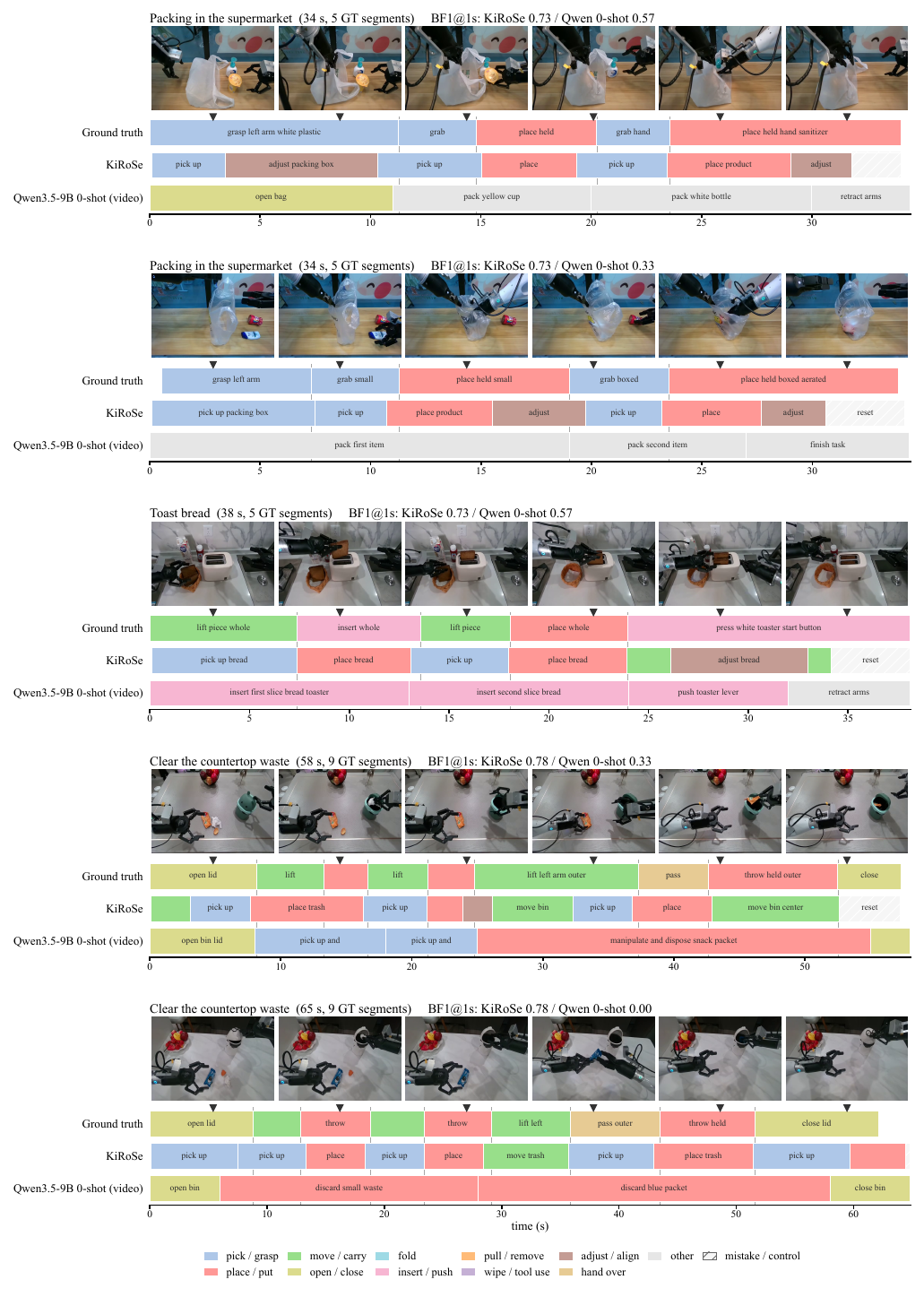}{\linewidth}
  \caption{\textbf{Segmentation examples on AgiBot World (zero-shot embodiment transfer).} Neither
  the motion tokenizer nor the reader sees AgiBot data, yet KiRoSe places boundaries at the
  grasp and release events and reaches Boundary F1 of 0.73--0.78 on all
  five episodes, above the video baseline on each (0.00--0.57). Its descriptions transfer at the level
  of actions (``pick up'', ``place''), while object names often follow YAM phrasing.}
  \label{fig:app_seg_agibot}
\end{figure}

\begin{figure}[p]
  \centering
  \appfig{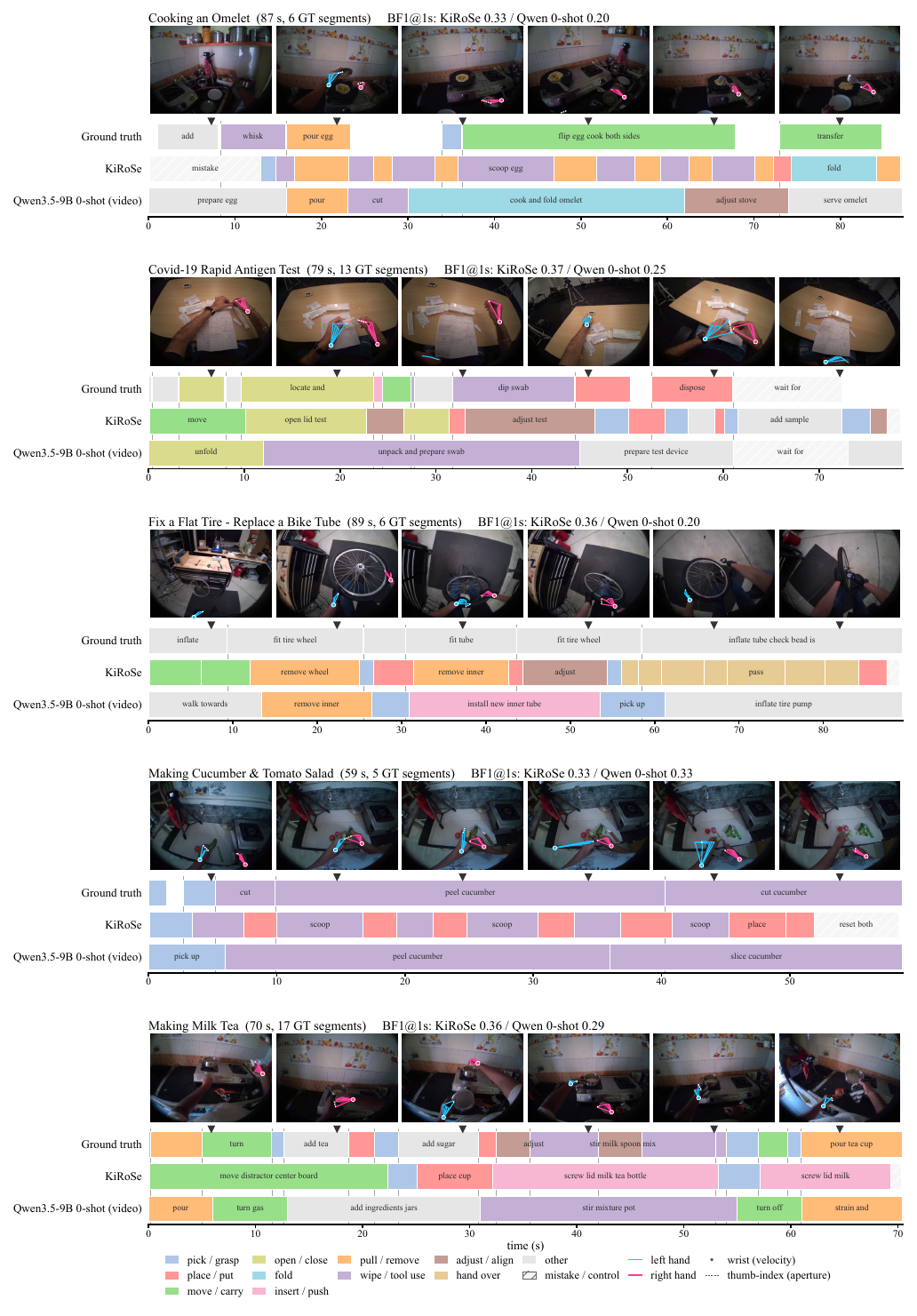}{0.94\linewidth}
  \caption{\textbf{Segmentation examples on Ego-Exo4D human-hand trajectories (zero-shot).}
  Keyframes are from the wearer's Aria camera (top 20\% cropped). Overlays show the hand joints
  that the conversion of Appendix~\ref{app:human} uses, projected from the 3D annotations into
  the image with an approximate equidistant fisheye model: the wrist (dot), which supplies linear velocity; the wrist-to-knuckle fan, which
  defines the palm frame and angular velocity; and the dotted thumb--index line, whose length
  supplies the aperture. Frames without an overlay for a hand have no valid 3D track at that
  instant. Although KiRoSe never sees human data and hands have no gripper, it matches or exceeds
  the video baseline on all five episodes; over all 114 trajectories with video, KiRoSe reaches
  0.375 Boundary F1 at $\pm1$\,s versus 0.255, and the gain depends on the motion tokens
  (Appendix~\ref{app:human}).}
  \label{fig:app_seg_human}
\end{figure}

Figures~\ref{fig:app_seg_yam}--\ref{fig:app_seg_human} show five trajectories from each
evaluation embodiment. Each panel shows six keyframes taken at the centers of six equal time bins
(triangles mark their timestamps), followed by the reference segmentation, the KiRoSe prediction,
and the prediction of the zero-shot Qwen3.5-9B video baseline for the same trajectory. The
keyframes are for the reader only; KiRoSe receives kinematics and the task instruction, never
images. Segments are colored by the class of their leading action verb, so matching colors
indicate agreement in \emph{what} is performed and aligned edges indicate agreement in \emph{when}
it changes; thin vertical lines mark reference boundaries. Hatched segments are control
annotations (\texttt{mistake}, \texttt{reset}), which are excluded from Grounded F1.

To avoid selecting favorable cases, examples are chosen by a fixed rule rather than by
inspection. Among episodes scored for both methods, we take the five tasks with the most
episodes and, within each task, the episode whose KiRoSe Boundary F1 at $\pm1$\,s is closest to
the task median; when fewer than five tasks are available (AgiBot has three), tasks are
revisited in the same order with the next-closest episode. The episode pools are the 190 YAM
held-out episodes with a Qwen prediction, the 357 AgiBot test and held-out-task episodes with a
Qwen prediction, and the 114 of the 193 human trajectories whose egocentric video is available
for the video baseline. The per-episode Boundary F1 of both methods is printed above each panel.
Under this fixed selection rule, KiRoSe matches or exceeds the video baseline on all 15 trajectories,
across three embodiments, two of which it never saw during training.

\begin{figure}[h!]
  \centering
  \appfig{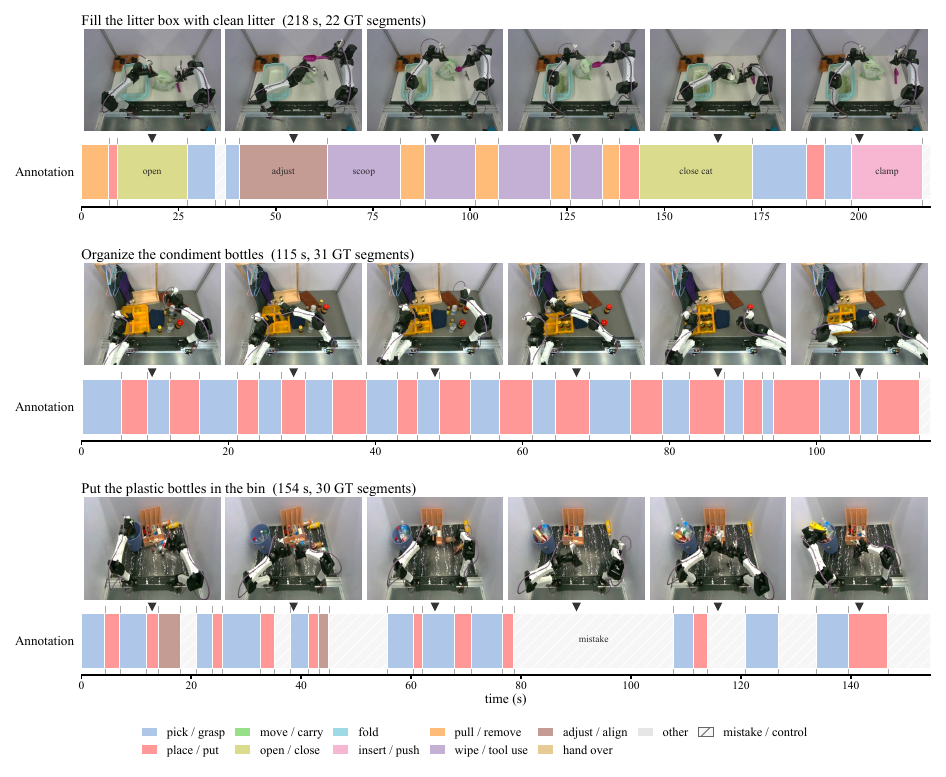}{0.78\linewidth}
  \caption{\textbf{Training annotations (YAM, ABC-130K).} Keyframes and the human subtask
  annotation of three training episodes, colored by leading verb as in
  Figures~\ref{fig:app_seg_yam}--\ref{fig:app_seg_human}. Annotations are dense and follow the
  arms' kinematic events: alternating grasp and place cycles when organizing bottles, open, scoop,
  and close phases when filling the litter box, and explicit \texttt{mistake} intervals (hatched)
  when a grasp fails.}
  \label{fig:app_seg_yam_train}
\end{figure}

For reference, Figure~\ref{fig:app_seg_yam_train} shows what the training supervision looks like.
Three YAM training episodes (the median-duration episode of each of the three tasks with the most
training episodes) are drawn with their annotation only. Annotators mark every grasp, release, and
tool-use event, so a two-minute episode carries 20--30 transitions. These transition times are the
only signal the kinematic encoder receives, and the paired descriptions supervise the reader.

\clearpage
\section{Kinematic Encoder, Transition Loss, and Tokenizer}
\label{app:encoder}

This appendix describes how the motion vocabulary is built (Figure~\ref{fig:overview}, left): a
dilated temporal encoder trained only to locate annotated transitions
(Sections~\ref{app:enc_arch}--\ref{app:enc_loss}), a $k$-means codebook fitted on its pooled
features (Section~\ref{app:enc_codebook}).

\subsection{Architecture}
\label{app:enc_arch}

\begin{figure}[t]
  \centering
  \appfig{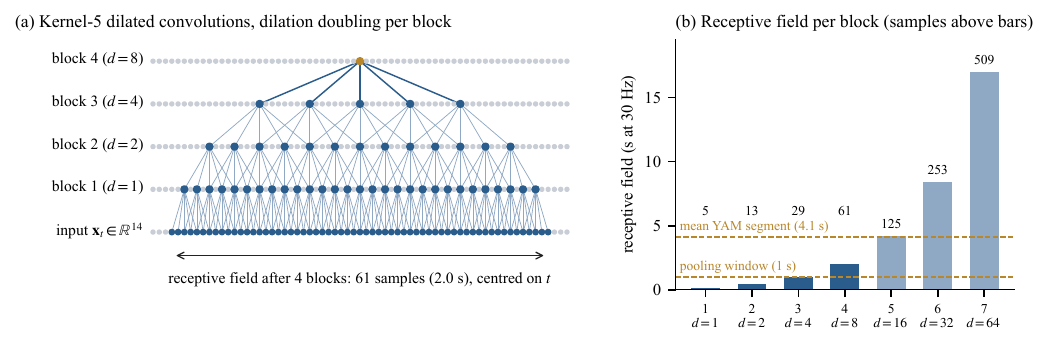}{\linewidth}
  \caption{\textbf{Dilated temporal convolutions in the kinematic encoder.} (a) Each residual block
  applies a kernel-5 convolution whose dilation doubles from block to block; tracing one output
  of the fourth block back to the input shows that the same five taps per layer cover 61 input
  samples. Padding is symmetric, so every feature sees both past and future motion. (b) With seven
  blocks (dilations $1,\ldots,64$), the receptive field grows to $1+4\sum_{i=0}^{6}2^i=509$ samples,
  or 17\,s at 30\,Hz. This spans several average-length YAM segments, so each feature can encode
  where the current motion sits relative to neighboring subtasks, while the parameter count grows
  only linearly with depth.}
  \label{fig:app_dilation}
\end{figure}

The encoder $f_\phi$ maps an episode $\mathbf{X}=(\mathbf{x}_1,\ldots,\mathbf{x}_T)$,
$\mathbf{x}_t\in\mathbb{R}^{14}$ at 30\,Hz, to features $\mathbf{H}=(\mathbf{h}_1,\ldots,\mathbf{h}_T)$,
$\mathbf{h}_t\in\mathbb{R}^{d}$ with $d=256$, in four steps (Table~\ref{tab:app_encoder}).

\paragraph{Per-episode standardization.} Each channel $c$ is z-scored with the statistics of its own
episode,
\begin{equation}
  \tilde x_{t,c}=\frac{x_{t,c}-\mu_c}{\max(\sigma_c,\,10^{-5})},\qquad
  \mu_c=\frac1T\sum_{t=1}^{T}x_{t,c},\quad
  \sigma_c^2=\frac1T\sum_{t=1}^{T}(x_{t,c}-\mu_c)^2,
  \label{eq:app_std}
\end{equation}
which removes differences in overall speed and gripper range between embodiments, at the cost of
making the encoder an offline model.

\paragraph{Input projection.} A pointwise ($1\times1$) convolution lifts the input to $d$ channels,
\begin{equation}
  \mathbf{h}^{(0)}_t=W_{\mathrm{in}}\tilde{\mathbf{x}}_t+\mathbf{b}_{\mathrm{in}},\qquad
  W_{\mathrm{in}}\in\mathbb{R}^{d\times 14}.
  \label{eq:app_proj}
\end{equation}

\paragraph{Dilated residual blocks.} For blocks $\ell=1,\ldots,L$ with $L=7$, kernel size $K=5$, and
dilation $\delta_\ell=2^{\ell-1}\in\{1,2,4,\ldots,64\}$, a centered dilated convolution aggregates the
previous layer at offsets that are multiples of $\delta_\ell$,
\begin{equation}
  \mathbf{u}^{(\ell)}_t=\sum_{k=-2}^{2}W^{(\ell)}_{k}\,\mathbf{h}^{(\ell-1)}_{t+k\delta_\ell}+\mathbf{b}^{(\ell)},
  \qquad W^{(\ell)}_k\in\mathbb{R}^{d\times d},
  \label{eq:app_dilconv}
\end{equation}
with zero padding at the episode ends, and the block adds a pointwise transform of it to its input,
\begin{equation}
  \mathbf{h}^{(\ell)}_t=\mathbf{h}^{(\ell-1)}_t+V^{(\ell)}\,\mathrm{Dropout}_{0.1}\!\big(\mathrm{GELU}(\mathbf{u}^{(\ell)}_t)\big)+\mathbf{c}^{(\ell)},
  \qquad V^{(\ell)}\in\mathbb{R}^{d\times d}.
  \label{eq:app_block}
\end{equation}
The output is $\mathbf{h}_t=\mathrm{LayerNorm}(\mathbf{h}^{(L)}_t)$, as in WaveNet-style temporal
convolution stacks~\citep{oord2016wavenet}.

\paragraph{Receptive field and size.} Because each block widens the context by $(K-1)\delta_\ell$
samples, $\mathbf{h}_t$ depends on the inputs
\begin{equation}
  R=1+(K-1)\sum_{\ell=1}^{L}\delta_\ell=1+4\,(2^{L}-1)=509\ \text{samples}\;(17.0\,\text{s at 30\,Hz}),
  \label{eq:app_rf}
\end{equation}
centered on $t$ (Figure~\ref{fig:app_dilation}). The context doubles with every block, whereas the
parameter count grows only linearly,
\begin{equation}
  |\phi|=\underbrace{(14d+d)}_{\text{Eq.~\ref{eq:app_proj}}}
  +L\,\underbrace{\big(Kd^2+d+d^2+d\big)}_{\text{Eqs.~\ref{eq:app_dilconv}--\ref{eq:app_block}}}
  +\underbrace{2d}_{\text{LayerNorm}}=2{,}760{,}448.
  \label{eq:app_params}
\end{equation}

\paragraph{Why dilation.}
Transitions are defined by their context: a gripper that stops closing marks the end of a grasp
only in light of the reaching motion that preceded it and the lift that follows. A stack of
undilated kernel-5 convolutions would need about 127 layers to reach a 17\,s context. Doubling
the dilation reaches it in 7 layers with 5 taps per layer, and the residual connections keep this
depth easy to optimize. Unlike attention, the convolution has no length limit and encodes
episodes of several minutes in a single pass.

\begin{table}[t]
\centering
\small
\caption{\textbf{Kinematic encoder, transition supervision, and tokenizer settings.}}
\label{tab:app_encoder}
\setlength{\tabcolsep}{4pt}
\begin{tabular}{@{}ll@{}}
\toprule
Input & $\mathbf{x}_t\in\mathbb{R}^{14}$ at 30\,Hz, per-episode z-scored \\
Projection & Conv1d($14\!\rightarrow\!256$, $k{=}1$) \\
Residual blocks & 7 $\times$ [Conv1d($256$, $k{=}5$, dilation $2^i$) $\rightarrow$ GELU $\rightarrow$ Dropout(0.1) $\rightarrow$ Conv1d($256$, $k{=}1$)] \\
Dilations & 1, 2, 4, 8, 16, 32, 64 (symmetric padding) \\
Output & LayerNorm(256); receptive field 509 samples (17.0\,s) \\
Encoder parameters & 2{,}760{,}448 \\
\midrule
Boundary head grid & 0.25\,s; box pools of half-width 0.25, 1, 4\,s, concatenated (768-d) \\
Boundary head & LN $\rightarrow$ Linear(768,256) $\rightarrow$ GELU $\rightarrow$ Conv1d(256, $k{=}5$, residual) $\rightarrow$ Linear(256,1) \\
Soft targets & $s(\tau)=\max_i\exp\!\big(-(\tau-\tau_i)^2/2\sigma^2\big)$, $\sigma=0.4$\,s \\
Loss & Focal BCE~\citep{lin2017focal}, $\gamma=2$, no class weighting \\
Optimizer & AdamW, lr $3\!\times\!10^{-4}$, weight decay 0.01, grad-norm clip 1.0 \\
Batch / epochs & 1 episode per step / 25 epochs, best validation loss \\
Data & 5{,}478 YAM training episodes; 608 validation episodes \\
Compute & $\approx$54 min on one GPU \\
\midrule
Pooling & 1\,s box average at a 0.5\,s stride (2 tokens/s), z-scored with training statistics \\
Codebook & $K=512$ $k$-means centroids fitted on 600k pooled YAM training vectors \\
\bottomrule
\end{tabular}
\end{table}

\subsection{Transition Loss}
\label{app:enc_loss}

\begin{figure}[t]
  \centering
  \appfig{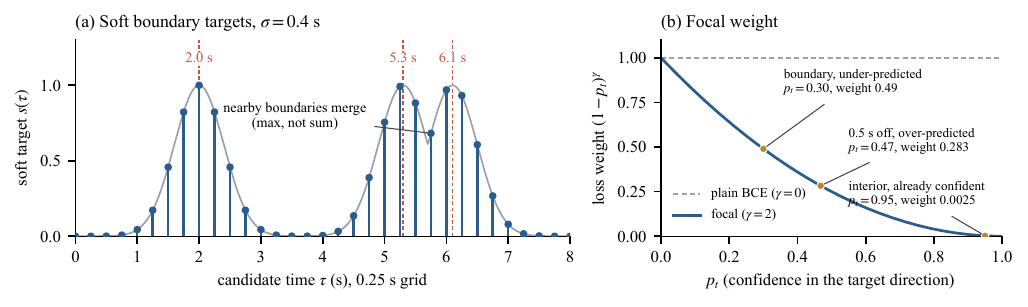}{\linewidth}
  \caption{\textbf{Transition-supervision loss.} (a) Soft targets on the 0.25\,s candidate grid for
  boundaries at 2.0, 5.3, and 6.1\,s ($\sigma=0.4$\,s). A grid point takes the height of the nearest
  Gaussian bump, so nearby boundaries merge instead of summing above one. (b) The focal weight
  $(1-p_t)^2$ against the model's confidence in the target direction, $p_t$. The three marked
  points are the worked examples of Table~\ref{tab:app_focal}: an already-confident interior point
  is down-weighted by a factor of 400, while an under-predicted boundary keeps half its weight.}
  \label{fig:app_loss}
\end{figure}

The encoder never sees a description; its only training signal is \emph{where} the annotated
boundaries are. The boundary head $b_\psi$ scores a uniform 0.25\,s candidate grid: at each
candidate, the encoder features are box-averaged over half-widths of 0.25, 1, and 4\,s, and the three
pooled vectors are concatenated and mixed by a small MLP, after which a residual kernel-5
convolution along the grid sharpens neighboring scores before a linear layer outputs the transition
logit. This is the implementation of the boundary head summarized in the main text:
the loss is evaluated on the candidate grid, not at every original 30\,Hz sample.
Writing $\mathbf{u}_\tau\in\mathbb{R}^{768}$ for the concatenated pooled vectors and
$\mathbf{U}=(\mathbf{u}_\tau)_{\tau\in\mathcal{T}}$, the head processes the grid sequence
and returns one logit per candidate. The loss has two parts (Figure~\ref{fig:app_loss}).

\paragraph{Soft targets.}
Each annotated boundary $\tau_i$ becomes a Gaussian bump, and a grid point takes the height of the
nearest bump,
\begin{equation}
  s(\tau)=\max_{i}\exp\!\Big(-\frac{(\tau-\tau_i)^2}{2\sigma^2}\Big),\qquad \sigma=0.4\,\text{s},
  \label{eq:app_soft}
\end{equation}
so that $s=1$ at the boundary, 0.82 at 0.25\,s, 0.46 at 0.5\,s, and 0.04 at
1\,s. Annotation timestamps are imprecise, and a hard 0/1 target would penalize a prediction that is
0.2\,s off as heavily as one that is 20\,s off.

\paragraph{Focal weighting.}
With the predicted transition probability $p_\tau=\sigma([b_\psi(\mathbf{U})]_\tau)$ at grid point
$\tau\in\mathcal{T}$ and its confidence in the target direction,
\begin{equation}
  p_{t,\tau}=s(\tau)\,p_\tau+\big(1-s(\tau)\big)(1-p_\tau),
\end{equation}
the transition loss of Section~\ref{sec:temporal_encoding} is
\begin{equation}
  \mathcal{L}_{\mathrm{transition}}=\frac{1}{|\mathcal{T}|}\sum_{\tau\in\mathcal{T}}
  \big(1-p_{t,\tau}\big)^{\gamma}\,
  \Big[-s(\tau)\log p_\tau-\big(1-s(\tau)\big)\log(1-p_\tau)\Big],\qquad \gamma=2.
  \label{eq:app_focal}
\end{equation}
Transitions are rare: YAM averages 9.2 boundaries per minute against 240 grid points, so only
about 4\% of grid points are the nearest point to a boundary. Plain BCE is dominated by easy interior
negatives and drives the head toward predicting no boundary anywhere; the focal factor down-weights the
loss from confident, correctly classified points (Table~\ref{tab:app_focal}). After
training, the head is discarded; the retained kinematic encoder has 2.76M parameters.

\begin{table}[h]
\centering
\small
\caption{\textbf{Worked focal-loss examples.} The gradient is carried by grid points near boundaries
that the model is still unsure about.}
\label{tab:app_focal}
\setlength{\tabcolsep}{5pt}
\begin{tabular}{@{}lccccc@{}}
\toprule
Grid point & $s$ & $p$ & BCE & $(1-p_t)^2$ & Contribution \\
\midrule
Interior, already confident & 0 & 0.05 & 0.051 & 0.0025 & 0.0001 \\
Boundary, under-predicted & 1 & 0.30 & 1.204 & 0.49 & 0.59 \\
0.5\,s from a boundary, over-predicted & 0.46 & 0.90 & 1.292 & 0.283 & 0.366 \\
\bottomrule
\end{tabular}
\end{table}

\subsection{Codebook}
\label{app:enc_codebook}

The frozen encoder's output is averaged over 1\,s windows centered on a 0.5\,s grid, producing two
256-dimensional vectors per second, which are z-scored with training-set statistics and assigned
to the nearest of $K=512$ centroids, $z_j=\arg\min_k\|\bar{\mathbf h}_j-\mathbf c_k\|_2^2$. The codebook is
fitted once, on 600k vectors sampled from the 1.90M pooled vectors of the YAM training episodes,
and never refitted; held-out episodes and other embodiments never enter the fit.

\section{Comparison with Motion Tokenizers}
\label{app:tokenizers}

Table~\ref{tab:main} compares SMT with three motion tokenizers, each read by the same kind of reader:
Qwen3.5-9B with rank-16 LoRA and new token rows, trained for one epoch on YAM segment annotations
and evaluated on the same 200 episodes from held-out tasks. This section gives the adaptation details
and token budgets.

\paragraph{Common setup.}
The reader, LoRA recipe, output format, decoding, and scorer are shared. The comparison therefore
evaluates complete tokenizer interfaces rather than isolating one design factor.
FAST, OAT, and MotionGPT are interpolated to 32\,Hz and scaled
with training-set 1st/99th percentiles and clipped to $[-1,1]$. Their readers are trained on 6{,}086
YAM episodes with the same answer-only JSON supervision, greedy decoding, and a 4{,}096-token output
limit. Empty or unparseable outputs are scored as predictions with no segments.

\paragraph{FAST~\citep{pertsch2025fast}.}
Each 4\,s chunk is transformed with an orthonormal temporal DCT; coefficients are scaled, rounded,
flattened frequency-major across channels, and compressed with byte-pair encoding (vocabulary
2{,}048). We use scale 0.5: at the default scale of 10, the longest training episode requires
102{,}125 tokens, beyond the reader's 32{,}768-token context.

\paragraph{OAT~\citep{liu2026oat}.}
Each 4\,s window ($128\times32$ samples) is mapped by a register transformer to eight 4-dimensional
registers, quantized by finite scalar quantization with levels $(8,5,5,5)$ (1{,}000 codes), and
decoded by a transformer trained with masked reconstruction loss and nested dropout over register
prefixes. Eight codes per 4\,s match SMT's rate of 2 tokens per second.

\paragraph{MotionGPT~\citep{jiang2023motiongpt}.}
A temporal CNN maps each 4\,s window to 32 vectors of dimension 512, quantized with a 512-entry
codebook (EMA updates and dead-code resets) and trained with masked Smooth-L1 reconstruction plus a
0.02 commitment loss, giving 8 codes per second.

\begin{table}[h]
\centering
\small
\caption{\textbf{Motion tokenizers as reader interfaces}, 200 YAM episodes from held-out tasks.
Motion tokens per second are pooled over episodes; reader input counts the whole prompt (instruction,
time markers, and motion tokens). Where shown, ranges are over three reader seeds. Reference: 24.3 segments per
episode.}
\label{tab:app_tokenizers}
\setlength{\tabcolsep}{3.5pt}
\begin{adjustbox}{max width=\linewidth}
\begin{tabular}{@{}llcccccc@{}}
\toprule
Tokenizer & Code trained for & Vocab. & Tokens/s & Reader input, mean (max) & BF1 $\pm$1\,s & BF1 $\pm$0.5\,/\,2\,s & Segments \\
\midrule
FAST & DCT + BPE compression & 2{,}048 & 9.5 & 2{,}313 (4{,}314) & 0.416 (0.401--0.431) & 0.218 / 0.502 & 59.8 \\
MotionGPT & reconstruction & 512 & 8.1 & 2{,}101 (4{,}265) & 0.412 & 0.227 / 0.499 & 57.1 \\
OAT & reconstruction & 1{,}000 & 2.0 & 1{,}162 (2{,}345) & 0.412 (0.403--0.420) & 0.221 / 0.512 & 52.7 \\
\textbf{SMT} & transition detection & 512 & 2.0 & 1{,}368 (2{,}836) & \textbf{0.630} & \textbf{0.477} / \textbf{0.710} & \textbf{28.0} \\
\bottomrule
\end{tabular}
\end{adjustbox}
\end{table}

\paragraph{Comparison.}
At 2 motion tokens per second, SMT exceeds OAT by 0.22 Boundary F1 at $\pm1$\,s and
0.26 at $\pm0.5$\,s. SMT uses 41\% and 35\% fewer reader input tokens than FAST and
MotionGPT, respectively (Table~\ref{tab:app_tokenizers}). The baseline readers predict
53--60 segments per episode against 24.3 in the reference, whereas SMT predicts 28.0.
These results compare complete tokenizer interfaces: input representations and reader training
populations differ, so they do not isolate the effect of the tokenizer objective alone.


\section{Human Hands as a Robot End Effector}
\label{app:human}

\begin{figure}[t]
  \centering
  \appfig{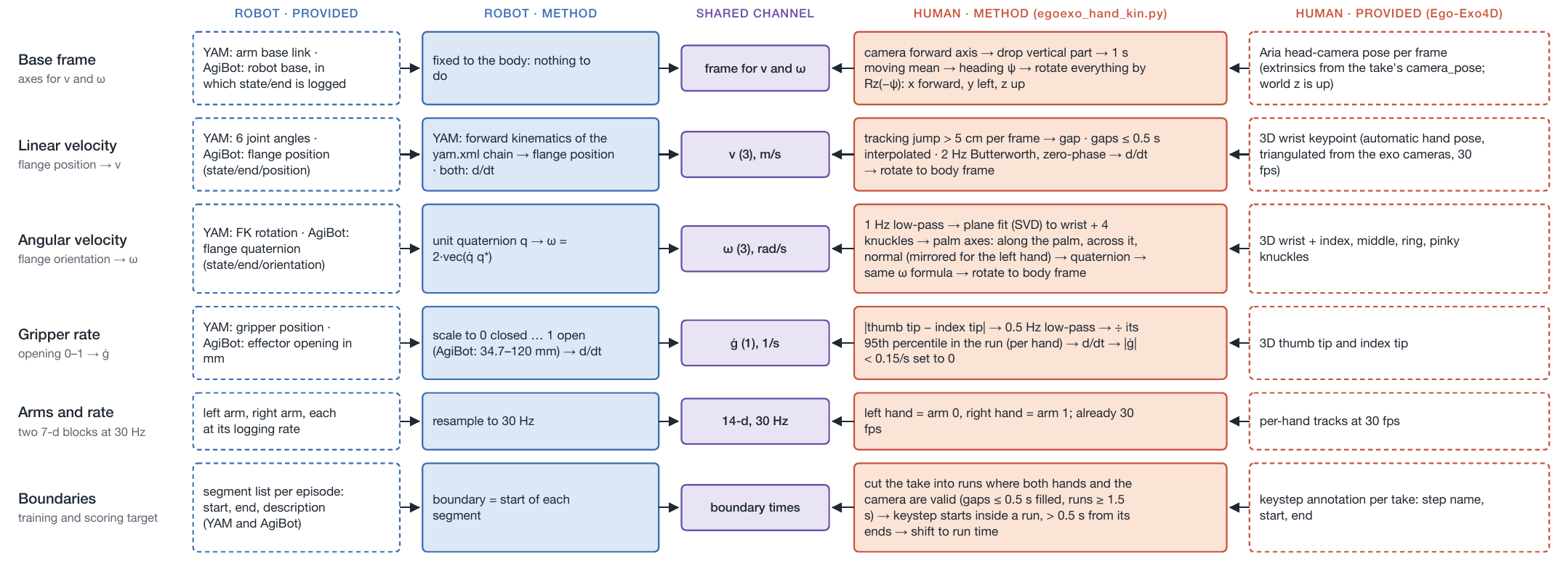}{\linewidth}
  \caption{\textbf{How each robot channel is built for a robot and for a human hand.} Rows are the
  channels of the shared representation (middle column). Dashed boxes: what each dataset provides;
  filled boxes: the conversion applied. Robots (left) need only forward kinematics, differentiation,
  and gripper normalization; human hands (right) require choosing a body frame, fitting a palm frame
  to the hand keypoints, using the thumb--index distance as a gripper, and projecting keystep
  annotations onto tracked intervals. The gap thresholds illustrated here describe the short-run
  conversion; the segmentation evaluation uses the longer-run protocol below.}
  \label{fig:app_hand_conversion}
\end{figure}

\begin{figure}[t]
  \centering
  \appfig{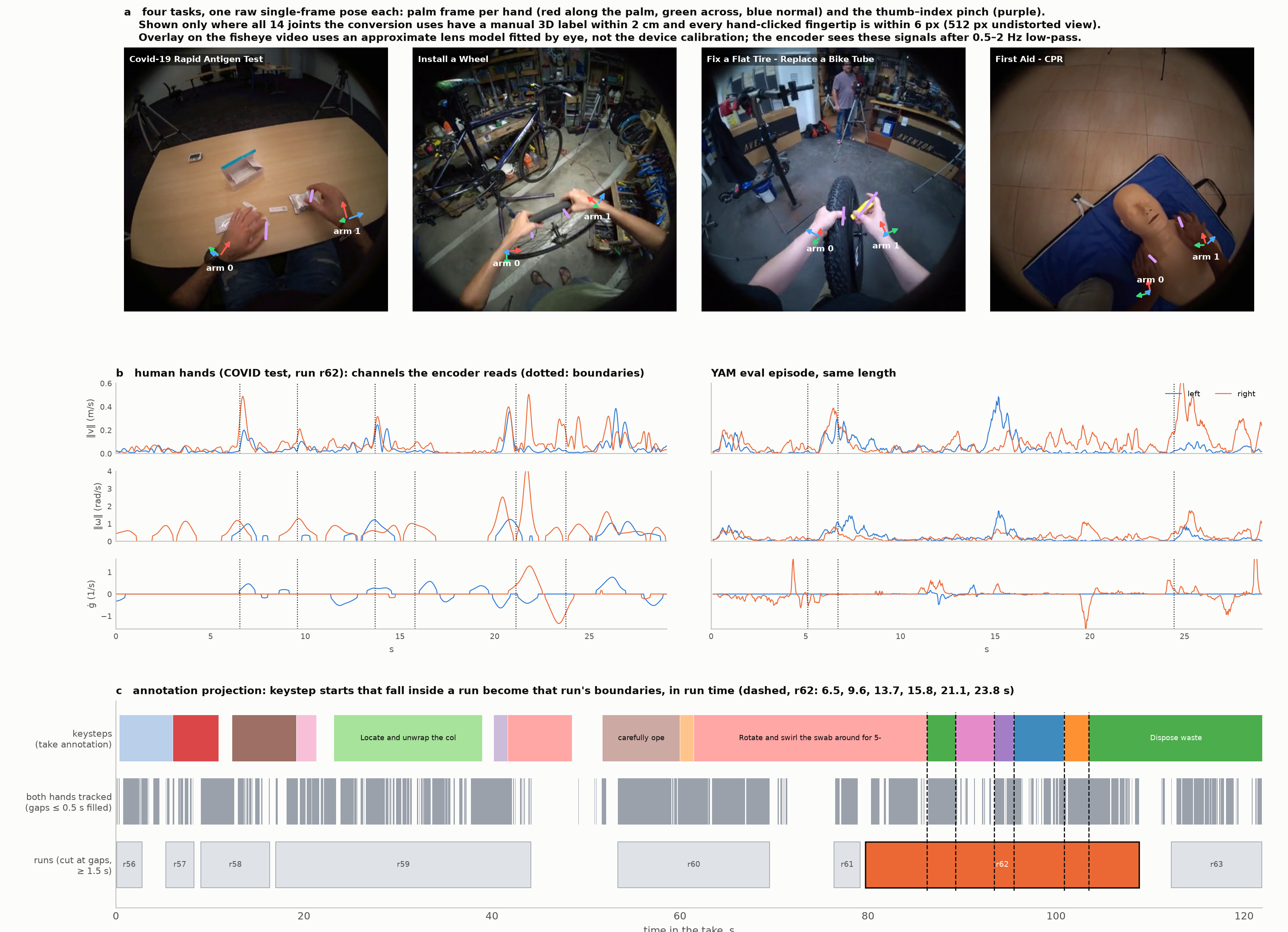}{\linewidth}
  \caption{\textbf{Human hands as a robot end effector, on real data.} (a) Four tasks, one
  single-frame pose each: the palm frame of each hand (red along the palm, green across, blue normal)
  and the thumb--index pinch (purple) that serves as the gripper, projected from 3D into the Aria
  image with an approximate equidistant fisheye model. Frames are shown only where all 14 joints the
  conversion uses agree with a manual 3D label within 2\,cm. (b) The resulting channels for a
  COVID-test run next to a YAM episode of the same length; dotted lines are annotated boundaries.
  Wrist speed and aperture rate rise at keystep boundaries much as robot speed and gripper rate do
  at subtask boundaries. (c) Annotation projection: keystep starts that fall inside a tracked run
  become that run's boundaries, in run time.}
  \label{fig:app_hand_alignment}
\end{figure}

All embodiments are mapped to the 14-dimensional bimanual representation of
Section~4.1,
\[
\mathbf{x}_t=\bigoplus_{h\in\{L,R\}}
[\mathbf{v}_{h,t}^{\top},\boldsymbol{\omega}_{h,t}^{\top},\dot g_{h,t}]^{\top},
\]
with the left hand or arm in channels
1--7 and the right in channels 8--14. Every stream is resampled by linear interpolation to a
uniform 30\,Hz grid. Figure~\ref{fig:app_hand_conversion} shows, channel by channel, what each
source provides and how it is converted, and Table~\ref{tab:app_projection} summarizes the result.
Both paths end in the same channels, so the encoder, tokenizer, and reader run unchanged on human
data. Figure~\ref{fig:app_hand_alignment} shows the conversion on real Ego-Exo4D data.

\begin{table}[h]
\centering
\small
\caption{\textbf{Construction of the shared 14-D representation per data source.}}
\label{tab:app_projection}
\setlength{\tabcolsep}{3pt}
\begin{tabularx}{\linewidth}{@{}l>{\raggedright\arraybackslash}X>{\raggedright\arraybackslash}X>{\raggedright\arraybackslash}X@{}}
\toprule
 & YAM (ABC-130K) & AgiBot G1 & Human hands (Ego-Exo4D) \\
\midrule
Pose source & Forward kinematics from 6 joint angles (MJCF chain of the YAM model) & Recorded end-effector position and orientation (\texttt{state/end}) & 3D hand keypoints (wrist, MCPs, thumb/index tips) \\
Frame & Per-arm base frame (z up, x forward) & Robot base frame (z up) & World frame, rotated by the smoothed head yaw \\
$\mathbf{v}$ & Central differences of position & Central differences of position & Central differences of 2\,Hz low-passed wrist position \\
$\boldsymbol{\omega}$ & From quaternion derivative & From quaternion derivative & From quaternion derivative of fitted palm frame \\
$g$ & Normalized gripper aperture (0 closed, 1 open) & Effector position, affinely mapped to $[0,1]$ and inverted & Thumb-tip--index-tip distance, normalized per trajectory \\
Native rate & Several hundred Hz & $\approx$30\,Hz & 30\,Hz (dense pseudo-ground truth) \\
Segments & Human subtask annotations & Human subtask annotations & Ego-Exo4D keystep annotations \\
\bottomrule
\end{tabularx}
\end{table}

\paragraph{Angular velocity.}
For all sources, orientation is expressed as a unit quaternion $q_t$. After enforcing hemisphere
continuity ($q_t\leftarrow -q_t$ if $\langle q_t,q_{t-1}\rangle<0$), angular velocity is computed as
$\boldsymbol{\omega}_t = 2\,\mathrm{vec}(\dot q_t \otimes q_t^{*})$, with $\dot q_t$ obtained by
central finite differences. We found this more reliable than differentiating rotation vectors,
which wrap discontinuously near $\pi$ for a substantial fraction of robot samples.

\paragraph{Robot grippers.}
YAM provides a normalized aperture that is interpolated onto the joint clock before
differentiation. The AgiBot effector reports a position in millimeters that \emph{increases} as
the gripper closes; we map it to an aperture $g=1-\mathrm{clip}((e-34.7)/(120-34.7),0,1)$ so
that, as for YAM, $g=1$ is open and $\dot g<0$ indicates closing.

\paragraph{Human source data.}
We use the Ego-Exo4D~\citep{grauman2024ego} ego-pose hand annotations, specifically the dense
automatic 3D keypoints, which are triangulated from the exocentric cameras at 30\,fps over the
full take, rather than the manual labels, which cover roughly 10\,s per take at 10\,Hz. We use
seven keypoints per hand: the wrist, the four metacarpophalangeal joints (index, middle, ring,
pinky), and the thumb and index fingertips. The head pose is taken from the extrinsics of the
wearer's Aria glasses.

\paragraph{Human coordinate frame.}
Keypoints are given in a gravity-aligned world frame (z up; the wrist lies below the head camera
in 99.7\% of 63{,}488 checked frames). Because the world frame's horizontal orientation is
arbitrary across takes, we rotate velocities about the vertical axis into a body-centric heading:
the camera's forward axis is projected onto the horizontal plane, smoothed with a 1\,s centered
moving average, and used to define yaw $\psi_t$; all vectors are rotated by $R_z(-\psi_t)$ so that
$x$ points forward and $y$ to the wearer's left, matching the robot base-frame convention. Only
yaw is removed, so gravity remains aligned with $z$ and head pitch or roll does not leak into the
hand motion.

\paragraph{Human wrist and palm.}
The linear velocity is the central difference of the wrist position after a zero-phase
second-order Butterworth low-pass filter (2\,Hz cutoff). The palm orientation, which plays the
role of the end-effector orientation, is fitted on a separately low-passed copy of the keypoints
(1\,Hz cutoff). Its first axis $\mathbf{e}_1$ points from the wrist to the centroid of the four
metacarpophalangeal joints. The palm normal $\mathbf{n}$ is the smallest singular vector of the
five palm points (wrist and four MCPs) about their centroid; its sign is fixed by
$\mathbf{e}_1\times(\mathbf{p}_{\text{index}}-\mathbf{p}_{\text{pinky}})$ for the right hand and the
mirrored product for the left hand, so both palms follow the same handedness convention. After
Gram--Schmidt orthogonalization of $\mathbf{n}$ against $\mathbf{e}_1$ and
$\mathbf{e}_2=\mathbf{n}\times\mathbf{e}_1$, the rotation $[\mathbf{e}_1,\mathbf{e}_2,\mathbf{n}]$ is
converted to a quaternion and differentiated as above. These filters matter: jitter in the
automatic tracks (about 4\,mm per frame) would otherwise appear as 0.1--0.2\,m/s of wrist speed and
over 1\,rad/s of palm rotation. The cutoffs trade temporal responsiveness for suppression of tracking noise.

\paragraph{Human grasp aperture.}
The thumb--index fingertip distance serves as the gripper aperture. It is low-passed at 0.5\,Hz
and divided by its 95th percentile within each trajectory and hand, so that an open hand is
$\approx1$ and a pinch is $\approx0$, as for the robot grippers. Opening rates with magnitude below
0.15\,s$^{-1}$ are set to zero.

\paragraph{Missing data and segmentation labels.}
Wrist jumps above 5\,cm between consecutive frames are treated as tracking failures and removed.
Segment labels are the Ego-Exo4D keystep annotations (start time, end time, step name), clipped
to each tracked run and shifted to run-relative time; keystep starts within 0.5\,s of either end
of a run are dropped (Figure~\ref{fig:app_hand_alignment}c).

\paragraph{Long trajectories.}
Cutting takes wherever tracking drops, with gaps of at most 0.5\,s interpolated, yields runs far
shorter than robot episodes (median 7\,s and 2 segments over 1{,}058 runs), most with a single
boundary. To match the temporal structure of the robot evaluations, the evaluation set instead
bridges tracking gaps of up to 2\,s by linear interpolation and keeps runs of at least 30\,s that
contain at least five keystep segments. This yields 193 trajectories from 12 tasks (median 64\,s
and 8 segments). About 46\% of their frames are interpolated, but 98\% of annotated boundaries
have real tracking within 0.5\,s. Interpolation nevertheless remains a limitation of this evaluation. Egocentric
video is available for 114 of the 193, which is the subset used for the video baseline.

\paragraph{The projection preserves segmentation-relevant motion.}
The converted human channels carry the same boundary evidence as robot channels: wrist speed rises
at keystep boundaries almost as much as end-effector speed does at YAM subtask boundaries (AUC 0.68
versus 0.72; Figure~\ref{fig:app_hand_alignment}b). KiRoSe, trained only on YAM, uses this evidence
zero-shot: on the 114 long trajectories with video, it reaches 0.375 Boundary F1 at $\pm1$\,s versus
0.255 for the zero-shot Qwen3.5-9B video model. The gain comes from the projected motion: removing
the motion tokens lowers KiRoSe to 0.296, and substituting another trajectory's tokens to 0.235.

\section{Progress Estimation}
\label{app:progress}

\paragraph{Frozen features.}
The progress probe reuses the transition-supervised encoder of Appendix~\ref{app:encoder}
with frozen weights; neither the encoder nor the codebook is fitted using progress, reversal, or
AgiBot data. The progress probe reads continuous features rather than discrete code indices. To make predictions online, the encoder runs on a causal window: at every 0.25\,s
step, it encodes the preceding 4\,s of kinematics (120 samples), standardized with fixed
channel statistics from the YAM training set rather than per episode. The probe also receives
an 8-dimensional state vector containing displacement accumulated since the subtask start:
$\int\mathbf{v}_h\,dt$ and $\int\dot g_h\,dt$ for each arm.

\paragraph{Progress head.}
The head is a per-step MLP,
\[
\mathrm{Linear}(264,256)\rightarrow\mathrm{GELU}\rightarrow\mathrm{Dropout}(0.1)\rightarrow\mathrm{Linear}(256,1),
\]
with 68{,}097 parameters and no temporal model beyond the frozen window features. It is trained on
4{,}000 YAM training subtasks and selected on 600 YAM validation subtasks; subtasks shorter
than 3.2\,s and control segments are excluded. 

\paragraph{Reversal construction.}
A subtask $[a,b)$ is reversed in place while the rest of the episode is unchanged. For
kinematics, the sampled sequence of all 14 channels inside $[a,b)$ is reversed in order and negated,
so that linear and angular velocities flip and gripper closing
becomes opening, which is the motion that would undo the subtask. For video, the subtask's
frames are played in reverse order.

\paragraph{Evaluation set.}
We use every non-control AgiBot subtask of at least 3.2\,s for which both video and kinematics are
available: 2{,}586 subtasks (median 5.6\,s) from 357 episodes. The episodes cover three tasks:
\emph{packing in the supermarket} and \emph{toast bread} from the test split (578 subtasks), and the
held-out task \emph{clear the countertop waste} (2{,}008 subtasks).

\paragraph{Baselines.}
ReWiND~\citep{rewind} does not release reward-model checkpoints, so we train it with the
authors' public code and recommended configuration (rewind augmentation, video subsampling,
gradient clipping, cosine schedule, batch size 1{,}024, 20 epochs), on MetaWorld only (MW)
and on the authors' preprocessed Open X-Embodiment embeddings (58{,}585 instructions) mixed with
MetaWorld demonstrations at the default ratio of 80\% Open-X to 20\% MetaWorld per batch (OXE). ReWiND uses
frozen DINOv2 ViT-B/14~\citep{oquab2024dinov2} frame embeddings and all-MiniLM-L12-v2~\citep{wang2020minilm}
text embeddings with a 13.3M-parameter reward transformer. Robometer-4B~\citep{robometer} is used
as released. Both baselines receive the subtask's frames at 5\,fps, resized to $224\times224$ and
subsampled to 16 frames, with the annotated subtask description as the instruction; the first
output is discarded, leaving 15 progress values per subtask. KiRoSe produces one value every
0.25\,s.

\paragraph{Metric.}
For each subtask, $\rho$ is the Spearman correlation between the predicted progress sequence and
its playback index; a constant prediction is assigned $\rho=0$. We report the mean over subtasks
and the fraction of reversed subtasks with $\rho<0$. Confidence intervals use an
episode-clustered bootstrap with 2{,}000 resamples.

\begin{table}[h]
\centering
\small
\caption{\textbf{Progress correlation by evaluation split}, forward\,/\,reversed mean Spearman
$\rho$. KiRoSe is the signed-rate head. Pooled over all 2{,}586 AgiBot subtasks, KiRoSe obtains
$+0.51\,[0.48,0.53]$ forward and $-0.68\,[-0.70,-0.67]$ reversed (95\% episode-clustered bootstrap
intervals), with $\rho<0$ on 86\% of reversed subtasks.}
\label{tab:app_progress_split}
\setlength{\tabcolsep}{4pt}
\begin{tabular}{@{}lccc@{}}
\toprule
Model & AgiBot test tasks & AgiBot held-out task & YAM held-out tasks \\
\midrule
ReWiND (MW)   & $+0.85$ / $+0.85$ & $+0.89$ / $+0.89$ & $+0.79$ / $+0.79$ \\
ReWiND (OXE)  & $+0.67$ / $+0.69$ & $+0.65$ / $+0.62$ & $+0.77$ / $+0.78$ \\
Robometer-4B  & $+0.68$ / $+0.08$ & $+0.72$ / $+0.05$ & $+0.53$ / $+0.30$ \\
KiRoSe        & $+0.50$ / $\mathbf{-0.75}$ & $+0.51$ / $\mathbf{-0.66}$ & $+0.64$ / $\mathbf{-0.66}$ \\
\bottomrule
\end{tabular}
\end{table}

Unless noted otherwise, KiRoSe results in this section use the final signed-rate probe. KiRoSe is
the only model whose progress rises during execution and falls under reversal on every split (Table~\ref{tab:app_progress_split}). Pooled over all 2{,}586 AgiBot subtasks, its progress
falls on 86\% of reversed subtasks, compared with 1\% and 4\% for the two ReWiND models and 46\% for
Robometer-4B. Neither the encoder nor the progress head was trained on AgiBot.

\section{Language Model Training}
\label{app:lm}

\paragraph{Vocabulary and parameterization.}
We start from Qwen3.5-9B~\citep{qwen35} in bfloat16 and add 512 special tokens
\texttt{<k0>}--\texttt{<k511>}, which extends the vocabulary from 248{,}077 to 248{,}589 entries. New input
embeddings are initialized to the mean of the existing rows plus Gaussian noise with 0.1 times
their standard deviation, and new output rows (the head is untied) to the mean of the existing
output rows. A gradient mask restricts updates of the embedding and output matrices to the new
rows; all other weights are adapted only through LoRA~\citep{hu2022lora}.

\paragraph{Prompt and target.}
The prompt uses the model's chat template with thinking disabled. Motion tokens are grouped by
second with a textual time marker, so the model reads absolute time directly instead of counting
tokens:

\begin{center}
\fbox{\begin{minipage}{0.95\linewidth}\footnotesize\ttfamily
Task: "place and organize the paper towels onto the shelf"\\
Robot arm motion as discrete codes, 2 per second, in time order:\\
\ [0s] <k93><k93> [1s] <k$\cdot$><k$\cdot$> [2s] <k$\cdot$><k$\cdot$> \ldots\ [190s] <k$\cdot$><k$\cdot$>\\
Segment this episode into subtasks. Respond with ONLY a JSON array of
\{"start","end","description"\} objects covering 0 to 190.5 seconds.
\end{minipage}}
\end{center}

The target is the reference segmentation serialized as a JSON array. Each segment contains
numeric \texttt{start} and \texttt{end} times in seconds and a \texttt{description} string.
The loss is applied to target tokens only. Prompts are not truncated. On the 200 held-out YAM
episodes, the SMT reader input averages 1,368 tokens and has a maximum of 2,836 tokens
(Table~\ref{tab:app_tokenizers}).

\begin{table}[h]
\centering
\small
\caption{\textbf{Reader training and inference settings.}}
\label{tab:app_lm}
\setlength{\tabcolsep}{4pt}
\begin{tabular}{@{}ll@{}}
\toprule
Base model & Qwen3.5-9B, bfloat16, gradient checkpointing \\
LoRA & rank 16, $\alpha=32$, dropout 0.05; q, k, v, o, gate, up, down projections \\
Trainable & LoRA adapters + 512 new input-embedding and output rows \\
Optimizer & AdamW, lr $2\!\times\!10^{-4}$ (constant, no warm-up), weight decay 0.01, grad-norm clip 1.0 \\
Batch & 1 episode per GPU, global batch 9 (3 GPUs, data parallel) \\
Data / epochs & 5{,}478 YAM training episodes / 1 epoch \\
Compute & $\approx$2.2\,h on 3 NVIDIA H200 GPUs \\
\midrule
Inference & LoRA merged; vLLM~\citep{kwon2023vllm}, greedy decoding, max 4{,}096 new tokens \\
Constraint & Logit bias $-100$ on all motion tokens to suppress their generation \\
Throughput & 200 YAM trajectories in 32\,s on one H200 (0.16\,s per trajectory, batched) \\
\bottomrule
\end{tabular}
\end{table}

\paragraph{Output parsing.}
Empty or unparseable responses are scored as predictions with no segments rather than removed,
so they count against the method.


\section{Reproducibility}
\label{app:repro}

\paragraph{Data and splits.}
The final encoder, codebook, reader, and progress head are fitted using only the 5,478 ABC-130K
YAM training episodes; the 608 YAM validation episodes are used for checkpoint selection.
The FAST, OAT, and MotionGPT baseline readers instead use all 6,086 YAM training and
validation episodes, excluding held-out tasks (Appendix~\ref{app:tokenizers}).
The YAM held-out-task evaluation contains 200 episodes from 10 tasks; the zero-shot Qwen
video comparison uses the 190 episodes with video. Thus the 200-episode YAM population
in the main segmentation table has a video-availability exception for Qwen.
The 2,653 AgiBot episodes summarized in the main dataset table are the held-out-task split.
The pooled AgiBot evaluation additionally includes 1,199 test episodes, giving 3,852 in total;
357 episodes across the two splits have Qwen video predictions.
Human segmentation uses 193 Ego-Exo4D trajectories from 12 tasks, of which 114 have video
(Appendix~\ref{app:human}). The progress evaluation uses 2,586 annotated subtasks from
357 AgiBot episodes (Appendix~\ref{app:progress}). None of the held-out-task, AgiBot, or human
evaluation episodes enters a fit.

\paragraph{Populations behind the main transfer comparisons.}
The KiRoSe bars in the main transfer figure use 200 YAM, 3,852 AgiBot, and 193 human
trajectories. The Qwen video bars use 190, 357, and 114 trajectories, respectively.
The Astra bars use smaller subsets of 50 YAM, 20 AgiBot, and 30 human trajectories;
they are not evaluations on the full populations of the neighboring KiRoSe bars.
The 20 AgiBot episodes comprise 10 test and 10 held-out-task episodes. The introduction's
AgiBot Boundary F1 comparison at $\pm0.5$\,s uses this matched 20-episode subset:
KiRoSe obtains 0.612 and Astra 0.301. KiRoSe's 0.40 in the transfer figure instead uses
the pooled AgiBot population. The action-level comparison also uses the matched 20 episodes;
the underlying scores are 0.593 for KiRoSe and 0.349 for Astra. These subset results should
be distinguished from full-population transfer results.

\paragraph{Motion-removal controls.}
The robot motion controls use the final reader on 200 held-out-task YAM episodes and the
pooled 3,852 AgiBot episodes. The instruction-only control removes motion tokens but retains
the instruction and time markers. The wrong-motion control substitutes a time-aligned token
sequence from another episode of the same task. At $\pm1$\,s, Boundary F1 on YAM is
0.630 with the correct motion, 0.349 with instruction only, and 0.342 with wrong motion;
on AgiBot the corresponding values are 0.614, 0.321, and 0.310.
The human controls in Appendix~\ref{app:human} use the 114-trajectory subset with video.

\paragraph{Compute.}
The encoder trains in about 54 minutes on one GPU. The $k$-means codebook is fitted once on the GPU.
The reader trains for one epoch with data parallelism on three NVIDIA H200 GPUs (global batch 9) in
about 2.2 hours; the progress head trains in under 10 seconds from cached features. Inference with
vLLM on one H200 processes 200 YAM trajectories in 32 seconds after model loading.

\paragraph{Determinism and variance.}
The encoder and codebook use seed 0, and inference uses greedy decoding. The final 14-D
reader was trained with one seed, so training-seed variability for that model is not measured.
The FAST and OAT reader ranges reported in Table~\ref{tab:app_tokenizers} span three seeds.

\paragraph{Evaluation.}
The segmentation results reported here use the same evaluator: greedy
one-to-one boundary matching within the stated tolerance, macro-averaged over episodes, and
one-to-one interval matching at IoU $\geq0.5$ with lexical ($\geq0.5$ token F1) or embedding
($\geq0.6$ MiniLM cosine) description agreement for Grounded F1, excluding control labels. Raw model
outputs, including failed parses, are retained.

\paragraph{Release.}
Upon publication, we will release the code for converting YAM, AgiBot, and Ego-Exo4D recordings into
the shared kinematic representation, the encoder and codebook checkpoints, the reader's LoRA adapter
and motion-token embeddings, the progress head, the evaluation splits and scorer, and the prompts used
for all baselines. Data derived from AgiBot World will be released under its CC BY-NC-SA 4.0 terms.
The Ego-Exo4D license does not permit redistributing the dataset, so instead of converted human data
we release the scripts that reproduce it from the official download.

\section{Data and Licenses}
\label{app:ethics}

\paragraph{Datasets.}
We use three existing datasets under their respective terms and collect no new data.
ABC-130K~\citep{abc130k} provides the YAM demonstrations and subtask annotations and is released under
the Apache License 2.0. AgiBot World~\citep{bu2025agibot} is used for evaluation only and is released
under CC BY-NC-SA 4.0, which permits our non-commercial research use. Ego-Exo4D~\citep{grauman2024ego}
is used for evaluation only, under the Ego-Exo4D license agreement, whose permitted purposes include
reproducing images from the dataset in research publications and which requires citing the dataset.
Its recordings were made with formal consent of the participants under the dataset's privacy and
ethics policies and the Project Aria Research Community Guidelines, in closed environments without
passersby. From Ego-Exo4D we use 3D hand keypoints and head-camera poses to construct the kinematic
input. Keystep annotations provide evaluation targets; they are not supplied to the
segmentation reader. Egocentric video frames appear only in figures and as input to the video
baselines.


\section{Broader Impact}
\label{app:impact}

\paragraph{Data curation and access.}
KiRoSe can provide initial segment boundaries and action descriptions for human review, supporting
demonstration retrieval, filtering, and inspection. It uses an open 9B-parameter reader and can
run locally on a single GPU without an API fee. Reported training and throughput measurements
are given in Appendix~\ref{app:repro}; we do not infer energy consumption from throughput alone.

\paragraph{Privacy.}
At inference, KiRoSe receives end-effector kinematics and the task instruction rather than images.
Local inference can avoid transmitting recordings to a third-party service. Kinematic data are
not an anonymity guarantee: motion patterns and task instructions can still disclose personal
information. Our human evaluation uses the consented research dataset described in
Appendix~\ref{app:ethics}.

\paragraph{Misuse: monitoring people.}
Because KiRoSe transfers to human hands, the same technique could be applied to wearable or
motion-capture data to segment and label what a person is doing, for example to monitor the pace or
activity of workers without their knowledge or consent. We evaluate human transfer only on a
consented research dataset, and we do not release models trained on human data. Deployments that
analyze human motion should obtain informed consent, limit collection to the stated purpose, and
follow applicable data-protection and workplace regulations. The kinematics-only design helps here as
well: it can be used without cameras, but it should not be treated as making human monitoring
acceptable where it would otherwise not be.

\paragraph{Safety of downstream use.}
Automatic segmentations can be wrong, and errors in labels used to train policies can propagate into
robot behavior. KiRoSe's accuracy is lower on human hands than on robots, and it has not been
evaluated on long-horizon mobile manipulation. Labels used for policy training should therefore be
spot-checked, particularly for embodiments and tasks far from the training distribution, and
automatic labels should not replace human review for safety-critical data such as failure and
recovery annotations.

\paragraph{Scope and bias.}
KiRoSe is trained on laboratory demonstrations from a single bimanual robot, with English annotations
written in one annotation style. Predictions can retain that phrasing on new embodiments, as
illustrated in Appendix~\ref{app:examples}, and the model may under-represent
activities, manipulation styles, or languages absent from the training data. Users applying it to new
domains should validate it on a labeled sample from that domain before relying on its output.

\end{document}